# Self-Adapting Noise-Contrastive Estimation for Energy-Based Models

Thesis Submitted to

**Tsinghua University**

in partial fulfillment of the requirement

for the degree of

**Master of Science**

in

**Computer Science and Technology**

by

**Nathaniel Xu**

Thesis Supervisor:    Professor Zhu Jun

**June, 2022**

# ABSTRACT

Training energy-based models (EBMs) with noise-contrastive estimation (NCE) is theoretically feasible but practically challenging. Effective learning requires the noise distribution to be approximately similar to the target distribution, especially for high-dimensional data. Previous work has explored the idea of modelling the noise distribution with a separate generative model, and then concurrently training this noise model with the EBM. While this method allows for more effective noise-contrastive estimation, it comes at the cost of extra memory and training complexity. Instead, this thesis proposes a self-adapting NCE algorithm which uses static instances of the EBM along its training trajectory as the noise distribution. During training, these static instances progressively converge to the target distribution, thereby circumventing the need to simultaneously train an auxiliary noise model. Moreover, we express this self-adapting NCE algorithm in the framework of Bregman divergences and show that it is a generalization of maximum likelihood learning for EBMs. The performance of our algorithm is evaluated across a range of noise update intervals, and experimental results show that shorter update intervals are conducive to higher synthesis quality.

# TABLE OF CONTENTS

# LIST OF FIGURES AND TABLES

# LIST OF SYMBOLS AND ACRONYMS

| | |
|---|---|
| pdf | probability density function |
| cdf | cumulative distribution function |
| $\nabla$ | gradient operator |
| $\nabla^2$ | Hessian operator |
| $p_{\text{data}}(\cdot)$ | data distribution |
| $p_{\text{n}}(\cdot)$ | noise distribution |
| $p_\theta(\cdot)$ | model distribution parameterized by some neural network $\theta$ |
| $\tilde{p}(\cdot)$ | unnormalized distribution |
| $\mathbf{x}_{-j}$ | the vector $\mathbf{x} = [x_1, ..., x_{j-1}, x_{j+1}, ...x_d]$ but with element $j$ excluded |
| $\odot$ | element-wise multiplication |
| $\hat{J}$ | unbiased estimator of $J$ |
| $\simeq$ | unbiased expression |
| $\langle \cdot, \cdot \rangle$ | inner product |
| $\| \cdot \|$ | L2 norm |
| $\text{Tr}(\cdot)$ | trace |
| $[\![\cdot]\!]$ | indicator function |
| $\sigma(\cdot)$ | sigmoid function |

# CHAPTER 1    INTRODUCTION

Creativity is a fundamental aspect of human intelligence and a key area of research for computer scientists. Current steps towards machine ingenuity focus on systems that can learn to generate new data based on previously observed data. These systems are based on a class of statistical models known as generative models, which attempt to learn either the complete distribution of observed data (explicit), or the process by which observed data is created (implicit).

Implicit density models, while extremely flexible with the types of distributions that can be modelled, do not directly model the underlying data distribution and are often more difficult to train and evaluate. For instance, generative adversarial networks (Goodfellow et al., 2014) are able to produce realistic images with state of the art quality (Karras et al., 2020). Despite this, training requires a careful balance between the generator and discriminator so that one does not overpower the other. Moreover, the min-max loss objective does not correlate with the generator's convergence and synthesis quality, making both training and evaluation more difficult (Salimans et al., 2016b).

Explicit density models directly model the data distribution (or an approximation of the data distribution) and therefore must be tractable. However, tractable models, while easy to train and evaluate, are often not as flexible. For example, autoregressive models assume that the data distribution can be factorized as a product of conditional distributions. Flow-based models assume that the data can be modelled as an invertible transform of a base distribution. Variational autoencoders assume that the data is inherently expressed as a particular latent variable model. In most cases, these assumptions can be too strong, resulting in generative models that are easy to train but unable to accurately learn the data distribution.

Energy-based models (EBMs), on the other hand, are a class of generative models that are compelling alternatives to those tractable (but inflexible) density models. Inspired by statistical physics, EBMs are characterized only by an energy function. This energy function is proportional to the logarithm of the unnormalized model-density, and thus completely describes the probability of given states (unlike implicit models). What makes EBMs so flexible is that, unlike other tractable density models, EBMs make no assumption that the model-density integrates to one. This gives the energy function the flexibility to take on any form as long as it is bounded; consequently, almost all continuous distributions can be modelled.

Because of their generality, simplicity, and flexibility, EBMs have found widespread applications in image generation (Du et al., 2019), text generation (Deng et al., 2020), reinforcement learning (Haarnoja et al., 2018), and even in discriminative learning (Grathwohl et al., 2020a). Despite the advantages, training EBMs remains challenging. Many divergence objectives require the normalized model-density, which is generally intractable. As such, improving the training of EBMs has been a point of emphasis in the academic community. Some recent examples of successful training algorithms include methods based on Score Matching (Song et al., 2019a), Stein Discrepancy (Grathwohl et al., 2020b), and Diffusion (Gao et al., 2021).

This thesis focuses on noise-contrastive estimation (NCE) (Gutmann et al., 2010), an algorithm that learns by contrasting true data with noisy data. Despite being discovered over a decade ago, training EBMs with NCE remains challenging; and the choice of noise distribution can have great influence over the performance of the algorithm, particularly for high-dimensional data. Ideally, the noise distribution should be close to the true data distribution; however, in practice the true data distribution is unknown. Recently, several attempts to scale NCE to high dimensions (Bose et al., 2018; Ceylan et al., 2018; Gao et al., 2020) have been proposed, but these methods still have unique drawbacks.

This study attempts to build on the previous work to scale NCE to high-dimensional problems. First, we analyze the improvements and limitations of NCE based training methods. Next, we propose a new training algorithm, and we qualitatively and quantitatively analyze its performance. Finally, several key insights are explained, including an unlikely connection between NCE and maximum likelihood learning. It should be noted that the methods discussed in this thesis are largely implemented with neural networks; as such, this thesis assumes a basic level of deep learning knowledge. A comprehensive review of deep learning is provided by (Goodfellow et al., 2016).

## 1.1 Contributions

The main contribution of this thesis is a novel EBM training algorithm inspired by noise-contrastive estimation. The emphasis of this work is not to achieve groundbreaking results, but rather to provide both empirical and theoretical insights about NCE. In particular, the thesis highlights the following:

1. **Self-adapting noise-contrastive estimation (AdaNCE)**: an improved NCE training algorithm where the noise distribution is taken to be static instances of the EBM along its training trajectory. This dynamic noise distribution progressively converges to the target distribution, thereby making NCE scalable to high-dimensional problems.

2. **Connection between AdaNCE and maximum likelihood estimation**: a surprising relationship between our proposed algorithm and maximum likelihood estimation is exposed in the context of EBM training.

3. **Generalizing AdaNCE under the framework of Bregman divergences**: self-adapting noise-contrastive estimation is explained in the framework of Bregman divergences. It turns out that this generalization also exhibits a connection to maximum likelihood estimation.

## 1.2 Thesis Outline

The main body of the thesis is divided into the following chapters:

**Chapter 2** is a review on all concepts used throughout this thesis. It covers a brief literature review of energy-based models and popular methods to train them, including noise-contrastive estimation. The basics of Markov chain Monte Carlo (MCMC) algorithms are also covered in this chapter, as MCMC sampling is a fundamental concept used in this work.

**Chapter 3** describes the AdaNCE algorithm in detail, its advantages over previous NCE based training methods, and practical considerations for improved training stability. The connection between AdaNCE and maximum likelihood estimation is also exposed.

**Chapter 4** explains NCE and AdaNCE as a minimization of Bregman divergences. Furthermore, we show that this generalization of AdaNCE also exhibits a connection to maximum likelihood estimation.

**Chapter 5** analyzes the performance of AdaNCE. The question of how generative models are evaluated is clarified, and the theory behind several popular evaluation metrics is explained. Experiment and parameter settings are given, as well as the results to all experiments.

**Chapter 6** concludes the thesis with a summary of findings and potential questions to be addressed in future work.

# CHAPTER 2  BACKGROUND

This chapter provides a brief review of key theoretical concepts used throughout the thesis. Part I covers Markov chain Monte Carlo methods, and part II covers energy-based models.

## 2.1  Part I: Markov Chain Monte Carlo

This section is all about Markov chain Monte Carlo (MCMC) methods, what they are, and why they are useful. We provide basic theory on Markov chains and stationary distributions, and explain four fundamental MCMC algorithms commonly encountered in the machine learning literature. For a more comprehensive and detailed tutorial on MCMC methods, readers are welcome to refer to Andrieu et al. (2003); Bishop (2006).

### 2.1.1  What is MCMC?

Suppose we have a complex distribution $p(x)$. Evaluation of this function is generally straightforward; however, sampling from $p(x)$ can pose significant challenges. If the distribution is simple enough (i.e., has a tractable and invertible cdf), we can rely on *inversion sampling* to sample a basic random variable and transform it into the desired distribution $p(x)$. For more complex distributions, we can apply *rejection sampling* to sample from a proposal distribution and reject unlikely samples. Unfortunately, both these sampling methods have critical drawbacks: inversion sampling is often inapplicable since in most cases the cdf and even the distribution itself are intractable; while rejection sampling is impractical as it fails to scale to high-dimensional distributions.

Markov chain Monte Carlo is a framework for sampling from a large class of distributions that scales well to high dimensions. Its strategy employs a Markov chain that explores the state space such that the chain converges to the distribution $p(x)$ from which we wish to sample. A particularly useful feature of this framework is that we are only required to evaluate $p(x)$ up to a normalizing constant. This makes MCMC methods perfect for sampling from intractable or unnormalized distributions.

### 2.1.2 Markov chains and stationary distributions

In order to understand Markov chain Monte Carlo, we must first understand what Markov chains are and the conditions in which they converge to the stationary distribution. A Markov chain is a stochastic process that describes a sequence of events, each of which only depends on the state of the previous event. For simplicity, we mainly consider discrete-time Markov chains; however, many of these properties can be extended to the corresponding continuous-time process.

A discrete-time Markov chain is a sequence of discrete random variables $\{X_1, ..., X_T\}$ that satisfies the following Markov property:

$$\mathbb{P}(X_n = x_n | X_1 = x_1, X_2 = x_2, ..., X_{n-1} = x_{n-1})$$
$$= \mathbb{P}(X_n = x_n | X_{n-1} = x_{n-1})$$
$$= \mathcal{T}_n(x_n | x_{n-1})$$

where $\mathcal{T}_n$ is the transition probability matrix at step $n$. This chain is *homogeneous* if the transition matrix is invariant to the step, i.e., $\mathcal{T}_n(x|z) = \mathcal{T}(x|z)$. Furthermore, the Markov process is fully described by its transition matrix $\mathcal{T}$ and initial distribution $p_1(x) := \mathbb{P}(X_1 = x)$, since the probability of any event $x$ at step $k$ can be expressed as:

$$p_k(x) = \sum_z \mathcal{T}(x|z) p_{k-1}(z).$$

What makes Markov chains remarkable is the existence of a class of distributions that are *stationary* (or *invariant*). This means that regardless of the initial distribution $p_1(\cdot)$, the chain will eventually stabilize at a stationary distribution $p^*(\cdot)$. It turns out that a distribution $p(\cdot)$ is invariant so long as the transition matrix $\mathcal{T}$ satisfies the following properties:

1. *Irreducible*: every state can be reached from every other state.

2. *Aperiodic*: the chain should not get trapped in cycles.

3. *Positive Recurrent*: the expected amount of time to return to the current state is finite.

A sufficient, but not necessary, condition for determining if $p(x)$ is a stationary distribution is the following *detailed balance* condition:

$$p(x)\mathcal{T}(x'|x) = p(x')\mathcal{T}(x|x'). \tag{2.1}$$

Therefore, all distributions that satisfy detailed balance on transition matrix $\mathcal{T}$ must also ensure that the underlying Markov chain is irreducible, aperiodic, and positive recurrent. Fortunately, it can be shown that homogeneous Markov chains are ergodic (i.e., irreducible, aperiodic and positive recurrent), subject only to weak restrictions on $p^*(x)$ and $\mathcal{T}$ (Neal, 1993) [①].

Many of the properties discussed for discrete-time Markov chains also apply in the continuous state-space. In particular, the continuous Markov chain is fully described by its transition kernel $\mathcal{K}$ and initial distribution $p_{t_0}(\cdot)$. The probability of any event $x$ at time $t_k$ is given by:

$$p_{t_k}(x) = \int_z \mathcal{K}_{t_k - t_{k-1}}(x|z) p_{t_{k-1}}(z) \, \mathrm{d}z,$$

where for homogeneous chains $\mathcal{K}_{t_k - t_{k-1}} \equiv \mathcal{K}$. The detailed balance condition for continuous Markov chains is identical to that of the discrete case (2.1), with the only difference being that the transition matrix $\mathcal{T}$ becomes the transition kernel $\mathcal{K}$.

### 2.1.3 MCMC methods

The core idea behind Markov chain Monte Carlo methods is to design a transition kernel $\mathcal{K}$ that ensures some target distribution $\pi(\mathbf{x})$ to be stationary. As such, any proposed Markov chain must guarantee that its kernel satisfies the detailed balance equation.

#### 2.1.3.1 Metropolis-Hastings

The Metropolis-Hastings (MH) algorithm (Hastings, 1970) is a classic MCMC method that many later sampling algorithms are based on. An MH step from the current state $\mathbf{x}$ involves first sampling a candidate $\mathbf{x}^*$ from a proposal distribution $q(\mathbf{x}^*|\mathbf{x})$, and then accepting the candidate with probability $\mathcal{A}(\mathbf{x}^*, \mathbf{x})$ where

$$\mathcal{A}(\mathbf{x}^*, \mathbf{x}) = \min\left(1, \frac{\pi(\mathbf{x}^*)q(\mathbf{x}|\mathbf{x}^*)}{\pi(\mathbf{x})q(\mathbf{x}^*|\mathbf{x})}\right). \tag{2.2}$$

The corresponding transition kernel $\mathcal{K}$ for this process is:

$$\mathcal{K}_{\mathrm{MH}}(\mathbf{x}^*|\mathbf{x}) = q(\mathbf{x}^*|\mathbf{x})\mathcal{A}(\mathbf{x}^*, \mathbf{x}) + \delta_{\mathbf{x}}(\mathbf{x}^*)r(\mathbf{x})$$

where $r(\mathbf{x}) = \int q(\mathbf{x}^*|\mathbf{x})(1 - \mathcal{A}(\mathbf{x}^*, \mathbf{x})) \, \mathrm{d}\mathbf{x}^*$ is the probability that any candidate state is rejected, and $\delta_{\mathbf{x}}(\cdot)$ is the impulse function centered at $\mathbf{x}$. We can show that this kernel satisfies detailed balance, and thus $\pi(\mathbf{x})$ is a stationary distribution of the Markov chain:

$$\begin{aligned}
\pi(\mathbf{x})\mathcal{K}_{\mathrm{MH}}(\mathbf{x}^*|\mathbf{x}) &= \pi(\mathbf{x})q(\mathbf{x}^*|\mathbf{x})\mathcal{A}(\mathbf{x}^*, \mathbf{x}) + \delta_{\mathbf{x}}(\mathbf{x}^*)\pi(\mathbf{x})r(\mathbf{x}) \\
&= \pi(\mathbf{x}^*)q(\mathbf{x}|\mathbf{x}^*)\mathcal{A}(\mathbf{x}, \mathbf{x}^*) + \delta_{\mathbf{x}^*}(\mathbf{x})\pi(\mathbf{x}^*)r(\mathbf{x}^*) \\
&= \pi(\mathbf{x}^*)\mathcal{K}_{\mathrm{MH}}(\mathbf{x}|\mathbf{x}^*).
\end{aligned}$$

---

[①] All Markov chains in this chapter are assumed to be ergodic.

The term $\delta_{\mathbf{x}}(\mathbf{x}^*)\pi(\mathbf{x})r(\mathbf{x})$ is equivalent to $\delta_{\mathbf{x}^*}(\mathbf{x})\pi(\mathbf{x}^*)r(\mathbf{x}^*)$ since the impulse function $\delta_{\mathbf{x}}(\mathbf{x}^*)$ is only nonzero at $\mathbf{x} = \mathbf{x}^*$. The equivalence of the remaining term is shown as follows:

$$
\begin{aligned}
\pi(\mathbf{x})q(\mathbf{x}^*|\mathbf{x})\mathcal{A}(\mathbf{x}^*, \mathbf{x}) &= \min\left(\pi(\mathbf{x})q(\mathbf{x}^*|\mathbf{x}), \pi(\mathbf{x}^*)q(\mathbf{x}|\mathbf{x}^*)\right) \\
&= \pi(\mathbf{x}^*)q(\mathbf{x}|\mathbf{x}^*)\min\left(\frac{\pi(\mathbf{x})q(\mathbf{x}^*|\mathbf{x})}{\pi(\mathbf{x}^*)q(\mathbf{x}|\mathbf{x}^*)}, 1\right) \\
&= \pi(\mathbf{x}^*)q(\mathbf{x}|\mathbf{x}^*)\mathcal{A}(\mathbf{x}, \mathbf{x}^*).
\end{aligned}
$$

Several properties of Metropolis-Hastings are worth highlighting. Firstly, the acceptance probability $\mathcal{A}$ does not require the target distribution to be normalized. This is because $\pi(\mathbf{x})$ appears on both the numerator and denominator, thereby allowing us to evaluate $\mathcal{A}$ using scaled versions of $\pi(\mathbf{x})$. Secondly, the choice of proposal distribution $q(\mathbf{x}^*|\mathbf{x})$ greatly influences the performance of the algorithm. If the proposal has small variance, the chain may not explore all modes of $\pi(\mathbf{x})$. If the proposal's variance is too high, the rejection rate may be very high, causing high correlations between samples and long convergence times. The Markov chain mixes well if all modes are visited and the acceptance rate is high. One common choice of proposal for the continuous sample space is a Gaussian centered on the current state.

### 2.1.3.2 Gibbs sampling

Gibbs sampling (Geman et al., 1984) is a special case of Metropolis-Hastings applied to joint distributions $\pi(x_1, ..., x_d)$. The basic idea is to sample each variable $x_i$ conditioned on the values of all other variables. This conditional distribution is referred to as the full-conditional and is given by $\pi(x_i|\mathbf{x}_{-i})$. The order in which variables are updated can be either deterministic or stochastic, so long as all variables are updated.

---

**Algorithm 2.1** Gibbs Sampler

---
1: Initialize $\mathbf{x}^{(1)}$
2: **for** $\tau = 1 : T$ **do**
3:     Sample $x_1^{(\tau+1)} \sim \pi(x_1|x_2^{(\tau)}, x_3^{(\tau)}, \ldots, x_d^{(\tau)})$
4:     Sample $x_2^{(\tau+1)} \sim \pi(x_2|x_1^{(\tau+1)}, x_3^{(\tau)}, \ldots, x_d^{(\tau)})$
5:     $\vdots$
6:     Sample $x_d^{(\tau+1)} \sim \pi(x_d|x_1^{(\tau+1)}, x_2^{(\tau+1)}, \ldots, x_{d-1}^{(\tau+1)})$
7: **end for**

---

We can view the Gibbs sampler as a Metropolis-Hastings algorithm with proposal distribution $q_j(\mathbf{x}^*|\mathbf{x}) = \pi(x_j^*|\mathbf{x}_{-j})\delta_{\mathbf{x}_{-j}}(\mathbf{x}_{-j}^*)$ for each of the $j = 1, \ldots, d$ variables. This

proposal ensures $\mathbf{x}^*_{-j} = \mathbf{x}_{-j}$ since these components have zero probability of being sampled. The factor that determines the corresponding acceptance probability is given by:

$$\mathcal{A}_j(\mathbf{x}^*, \mathbf{x}) = \min\left(1, \frac{\pi(\mathbf{x}^*)q_j(\mathbf{x}|\mathbf{x}^*)}{\pi(\mathbf{x})q_j(\mathbf{x}^*|\mathbf{x})}\right)$$

$$= \min\left(1, \frac{\pi(x_j^*|\mathbf{x}^*_{-j})\pi(\mathbf{x}^*_{-j})\pi(x_j|\mathbf{x}^*_{-j})\delta_{\mathbf{x}^*_{-j}}(\mathbf{x}_{-j})}{\pi(x_j|\mathbf{x}_{-j})\pi(\mathbf{x}_{-j})\pi(x_j^*|\mathbf{x}_{-j})\delta_{\mathbf{x}_{-j}}(\mathbf{x}^*_{-j})}\right) = 1,$$

where the ratio term equates to one since $\mathbf{x}^*_{-j} = \mathbf{x}_{-j}$. Therefore, the Metropolis-Hastings steps are always accepted. The biggest obstacle for Gibbs sampling is being able to efficiently sample from all full-conditionals of the distribution. In most cases, these conditional probabilities are either unknown or intractable. This is why Gibbs sampling is more commonly seen in directed graphical models, whose conditionals can be explicitly set to be simple distributions like Gaussian.

### 2.1.3.3 Langevin Monte Carlo

The Metropolis-adjusted Langevin algorithm (MALA) (Besag, 1994) is a gradient based MCMC sampling algorithm inspired by an (over-damped) Langevin diffusion process. Candidate states are proposed through Langevin dynamics modelled by the following stochastic differential equation (SDE):

$$\frac{\partial \mathbf{x}}{\partial t} = \frac{1}{2}\nabla_{\mathbf{x}}\log\pi(\mathbf{x}) + \frac{\partial \mathbf{w}}{\partial t}$$

where $\mathbf{w}$ is the standard Brownian motion. The solution to this SDE is a Markov process which, under some mild assumptions, has stationary distribution $\pi(\mathbf{x})$. This means that if we could simulate $\mathbf{x}_t$ along its diffusion path $t \to \infty$, the distribution $\mathbb{P}(\mathbf{x}_t \in \cdot)$ would converge to $\pi(\mathbf{x})$.

One way we can approximate this sample path is by discretizing the time steps into intervals of $\tau$. This numerical method is known as the *Euler–Maruyama approximation* and can be viewed as an extension of the Euler method for ordinary differential equations (ODEs). At step $k$ of the approximation, the Langevin diffusion is governed by the following expression:

$$\frac{\mathbf{x}_{k+1} - \mathbf{x}_k}{\tau} = \frac{1}{2}\nabla_{\mathbf{x}}\log\pi(\mathbf{x}_k) + \frac{\mathbf{w}_{k+1} - \mathbf{w}_k}{\tau}$$

$$\mathbf{x}_{k+1} = \mathbf{x}_k + \frac{\tau}{2}\nabla_{\mathbf{x}}\log\pi(\mathbf{x}_k) + (\mathbf{w}_{k+1} - \mathbf{w}_k)$$

$$\mathbf{x}_{k+1} = \mathbf{x}_k + \frac{\tau}{2}\nabla_{\mathbf{x}}\log\pi(\mathbf{x}_k) + \sqrt{\tau}\,\boldsymbol{\epsilon} \quad \text{where } \boldsymbol{\epsilon} \sim \mathcal{N}(\mathbf{0}, I), \qquad (2.3)$$

through which the definition of Brownian motion implies $\mathbf{w}_{k+1} - \mathbf{w}_k \sim \mathcal{N}(\mathbf{0}, \tau I)$. As $\tau \rightarrow 0$ and $k \rightarrow \infty$, the discretization error becomes zero and the approximation $\mathbf{x}_t$ converges to samples from the stationary distribution $\pi(\mathbf{x})$. In reality, $\tau$ is nonzero; and so to correct for the discretization error, we can use a Metropolis-Hastings step to ensure that the chain converges to $\pi(\mathbf{x})$. The proposal distribution for this MH step is directly inferred from the discretized Langevin dynamics to be:

$$q(\mathbf{x}^*|\mathbf{x}) = \mathcal{N}(\mathbf{x} + \frac{\tau}{2}\nabla_{\mathbf{x}}\log\pi(\mathbf{x}), \tau I).$$

It turns out that Langevin Monte Carlo is once again a special case of Metropolis-Hastings, where the proposed states are determined through a process resembling stochastic gradient descent with noise.

### 2.1.3.4 Hamiltonian Monte Carlo

Hamiltonian dynamics provides a framework for how particles move in a closed system. For example, it can be used to model the orbit of planets around the sun, or the motion of a golf ball rolling down a field. In particular, a particle's energy under Hamiltonian dynamics is completely described by the *Hamiltonian* $H(\mathbf{x}, \mathbf{v})$ as a function of position $\mathbf{x}$ and momentum $\mathbf{v}$:

$$H(\mathbf{x}, \mathbf{v}) = U(\mathbf{x}) + K(\mathbf{v}) = -\log\pi(\mathbf{x}) + \frac{1}{2}\mathbf{v}^{\mathsf{T}}M^{-1}\mathbf{v}$$

where $U(\mathbf{x})$ and $K(\mathbf{v})$ are the potential and kinetic energies, respectively, and $M$ is a symmetric positive definite mass matrix. At any time $t$, the motion of the system (i.e., $(\mathbf{x}(t), \mathbf{v}(t))$) is explained by *Hamilton's equations*:

$$\begin{aligned} \frac{\mathrm{d}x_i}{\mathrm{d}t} &= \frac{\partial H(\mathbf{x}, \mathbf{v})}{\partial v_i} & & \nabla_t\mathbf{x}(t) = M^{-1}\mathbf{v} \\ \frac{\mathrm{d}v_i}{\mathrm{d}t} &= -\frac{\partial H(\mathbf{x}, \mathbf{v})}{\partial x_i} & \implies & \nabla_t\mathbf{v}(t) = -\nabla_{\mathbf{x}}U(\mathbf{x}), \end{aligned}$$

where the equations on the right side are Hamilton's equations if we define the kinetic and potential energies to be $\frac{1}{2}\mathbf{v}^{\mathsf{T}}M\mathbf{v}$ and $-\log\pi(\mathbf{x})$, respectively.

The naive way of simulating Hamiltonian dynamics is with Euler's method, which discretizes the time steps into $\epsilon$. The resulting approximate position and momentum vectors at time $t$ are given by:

$$\mathbf{v}(t + \epsilon) = \mathbf{v}(t) + \epsilon\nabla_t\mathbf{v}(t) = \mathbf{v}(t) - \epsilon\nabla_{\mathbf{x}}U(\mathbf{x}(t))$$

$$\mathbf{x}(t + \epsilon) = \mathbf{x}(t) + \epsilon\nabla_t\mathbf{x}(t) = \mathbf{x}(t) + \epsilon M^{-1}\mathbf{v}(t).$$

Unfortunately, it turns out that the standard Euler method is not very good at preserving volume in systems of differential equations, meaning that the trajectory often diverges for oscillatory motion. A stronger alternative is the *leapfrog integrator*, which first performs half a step for momentum, and then a full step for position, followed by half a step for momentum while using the new position variable. The update steps are given below, where for readability we use subscripts to denote the time index:

$$\mathbf{v}_{t+\frac{\epsilon}{2}} = \mathbf{v}_t - \frac{\epsilon}{2}\nabla_{\mathbf{x}}U(\mathbf{x}_t)$$

$$\mathbf{x}_{t+\epsilon} = \mathbf{x}_t + \epsilon M^{-1}\mathbf{v}_{t+\frac{\epsilon}{2}}$$

$$\mathbf{v}_{t+\epsilon} = \mathbf{v}_{t+\frac{\epsilon}{2}} - \frac{\epsilon}{2}\nabla_{\mathbf{x}}U(\mathbf{x}_{t+\epsilon}).$$

One useful observation of the leapfrog method is that we can combine the first step of the current cycle with the last step of the previous cycle. As such, only one gradient needs to be computed per iteration. Furthermore, unlike Euler's method, the leapfrog method preserves volume exactly.

Hamiltonian Monte Carlo (HMC) (Neal, 2011) is a MCMC method, in which candidate samples are first proposed using Hamiltonian dynamics. This is done by running a leapfrog integrator for $L$ steps, where the momentum vector $\mathbf{v}$ is sampled from the Boltzmann distribution corresponding to the kinetic energy (typically Gaussian). After each simulation, the proposed samples $\mathbf{x}^*$ are accepted with probability $A(\mathbf{x}^*, \mathbf{x}) = \min\left(1, e^{-H(\mathbf{x}^*, \mathbf{v}^*)+H(\mathbf{x}, \mathbf{v})}\right)$.

---

**Algorithm 2.2** Hamiltonian Monte Carlo

1: Initialize $\mathbf{x}^{(0)}$
2: **for** $i = 1 : N$ **do**
3:     $u \sim \mathcal{U}(0, 1)$
4:     $\mathbf{v} \sim \mathcal{N}(\mathbf{0}, M)$
5:     $\mathbf{x}^*, \mathbf{v}^* = \text{Hamiltonian-Dynamics}(\mathbf{x}^{(i-1)}, \mathbf{v}, L, \epsilon)$
6:     **if** $u < \min\left(1, e^{-H(\mathbf{x}^*, \mathbf{v}^*)+H(\mathbf{x}, \mathbf{v})}\right)$ **then**
7:         $\mathbf{x}^{(i)} = \mathbf{x}^*$
8:     **else**
9:         $\mathbf{x}^{(i)} = \mathbf{x}^{(i-1)}$
10:     **end if**
11: **end for**

---

In theory, instances along any Hamiltonian trajectory should be energy preserving and thus $H(\mathbf{x}^*, \mathbf{v}^*) = H(\mathbf{x}, \mathbf{v})$. However, in reality, this trajectory is discretized and the energies are not conserved. As such, the Metropolis adjust step is needed to reject trajectories that have deviated too far. It is also worth mentioning that the stationary distribution $\pi(\mathbf{x})$ again does not need to be normalized, since $-H(\mathbf{x}*, \mathbf{v}^*) + H(\mathbf{x}, \mathbf{v})$ cancels out any scalar factor of $\pi(\cdot)$. The proof that HMC indeed converges to $\pi(\mathbf{x})$ is not within the scope of this thesis, and interested readers may refer to (Neal, 2011; Betancourt, 2017).

## 2.2 Part II: Energy-Based Models

This section provides a general overview of energy-based models and how to train them. First, we explain what EBMs are and why they are useful. Then, several different architectures are illustrated, starting from models inspired by statistical physics to models based on deep neural networks. Finally, we cover common methods for training EBMs and their advantages and disadvantages. Readers who are interested in a more comprehensive tutorial on EBMs should refer to (LeCun et al., 2006; Song et al., 2021a).

### 2.2.1 What are energy-based models?

An energy-based model is a Boltzmann distribution on random variable $\mathbf{x}$ described by some energy potential $E(\mathbf{x})$ and temperature $T$. If we assume the model is parameterized by $\theta$, then the distribution becomes:

$$p_\theta(\mathbf{x}) = \frac{1}{Z_\theta} e^{-\frac{1}{T} E_\theta(\mathbf{x})}$$

where $Z_\theta$ is the often intractable normalization constant (partition function) that satisfies:

$$Z_\theta = \int e^{-\frac{1}{T} E_\theta(\mathbf{x})} \, \mathrm{d}\mathbf{x}.$$

In practice, the temperature variable is often disregarded as it can be absorbed into the energy function, allowing $E_\theta(\mathbf{x})$ to sufficiently describe the model.

The Boltzmann distribution establishes an important relationship between energy and probability: lower energy states have higher probabilities. Therefore, learning a data distribution translates to learning an energy landscape where frequently occurring states have lower energies. Particularly noteworthy is that since EBMs are completely described by their scalar energy functions $E(\mathbf{x}) : \mathcal{X} \to \mathbb{R}$, the density estimation problem reduces to a nonlinear regression problem.

The key advantage that EBMs have over tractable density models is flexibility: EBMs are able to model any positive continuous distribution. This flexibility is attributed to the fact that $E_\theta(\mathbf{x})$ can describe any bounded function, and that the energy is not required to be normalized as probabilities. For example, the multivariate Gaussian distribution can be expressed as a Boltzmann distribution with energy class $E(\mathbf{x}) = \frac{1}{2}(\mathbf{x} - \mu)^\mathsf{T} \Sigma^{-1}(\mathbf{x} - \mu) + C$;

and in fact, the energies of any positive density $p(\mathbf{x})$ is simply $E(\mathbf{x}) = -\log p(\mathbf{x}) + C$, where $C$ is any constant. Another advantage is that EBMs provide a general framework for directed graphical models. So far we only considered the case where $x$ is independent; however, it is straightforward to extend the model to dependent variables by considering the conditional energy function $E(\mathbf{x}|\mathbf{z})$. In this manner, EBMs are able to learn any type of structured data.

### 2.2.2 Architectures: past to present

Before the explosion of deep learning and neural networks, early energy-based models were described using simple energy functions on collections of binary-valued atoms. The *Hopfield network* (Hopfield, 1982) is one of the earliest such models, whose energy function depends on the state of each atom and the pairwise interactions between them. If the states of $n$ atoms are described by the vector $\mathbf{x} = (x_1, x_2, ..., x_n)$ where $x_i \in \{-1, +1\}$, then the energy of the Hopfield network is:

$$E(\mathbf{x}) = -\frac{1}{2}\sum_{i,j} w_{i,j} x_i x_j - \sum_i b_i x_i$$
$$= -\left(\frac{1}{2}\mathbf{x}^\mathsf{T}\mathsf{W}\mathbf{x} + \mathbf{b}^\mathsf{T}\mathbf{x}\right),$$

where $\mathsf{W}$ is a symmetric weight matrix with diagonal zero and $\mathbf{b}$ is a bias vector. Hopfield networks belong to a more general class of models known as *Ising models* (Ising, 1925), in which the weight matrix $W$ does not need to be symmetric. The primary application of Hopfield networks is in associative memory, otherwise known as pattern completion or image inpainting. The basic idea is to train the network on a collection of bit vectors. Once trained, corrupted patterns with missing bits are presented to the network and the uncorrupted bits are estimated.

So far, Hopfield networks are only able to capture dependency between pairs of atoms, limiting the overall modelling power of the networks. One way to increase the model complexity is by including triple-wise interactions into the energy function; however, this makes inference much more challenging. A better way is adding a layer of hidden atoms to the model that interacts with the original visible atoms and themselves. The resulting network is known as an unrestricted *Boltzmann machine*, and is a generalization of Hopfield and Ising models; as such, the energy function for Boltzmann machines remains the same. Particularly noteworthy is that unrestricted Boltzmann machines are universal approximators for any discrete probability distribution (Le Roux et al., 2008).

Unfortunately, the added complexity with Boltzmann machines makes exact inference intractable and approximate inference (e.g., using Gibbs sampling) slow. To speed up approximate inference, the *restricted Boltzmann machine* (RBM) (Smolensky, 1986; Hinton et al., 2006) assumes that all intralayer connections are removed, resulting in a bipartite graph of hidden and visible atoms. The resulting energy function is:

$$E(\mathbf{v}, \mathbf{h}) = -\sum_{i,j} w_{i,j} v_i h_j - \sum_i a_i v_i - \sum_j b_j h_j$$

$$= -\left(\mathbf{v}^\mathsf{T} \mathbf{W} \mathbf{h} + \mathbf{a}^\mathsf{T} \mathbf{v} + \mathbf{b}^\mathsf{T} \mathbf{h}\right),$$

where $\{\mathbf{v}, \mathbf{h}\}$ is the set of visible and hidden atoms and $\{\mathbf{a}, \mathbf{b}\}$ is the corresponding bias. Although not as powerful as Boltzmann machines, RBMs allow for efficient gradient-based training algorithms like Contrastive Divergence (CD) (Hinton, 2002). This improved training has allowed RBMs to find success in tasks like dimension reduction (Hinton et al., 2006), classification (Larochelle et al., 2008), and topic modelling (Salakhutdinov et al.).

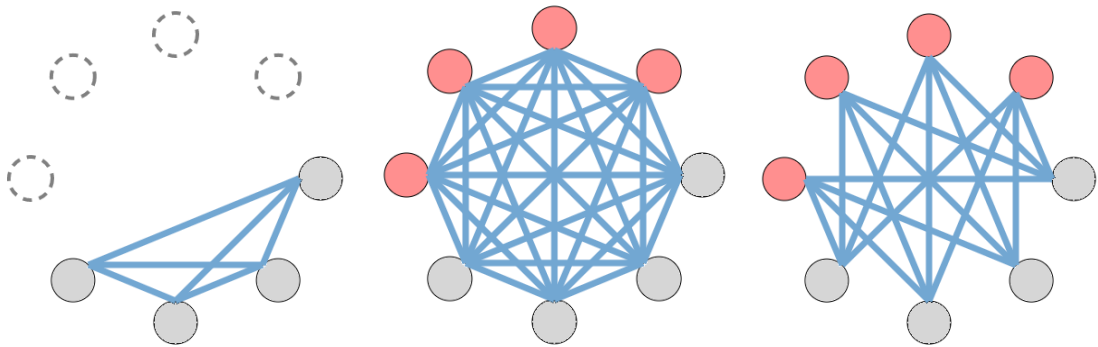

**Hopfield network**: All pairwise interactions between visible (input) nodes are assigned a weight

**Boltzmann machine**: All pairwise interactions between visible and hidden nodes are assigned a weight

**Restricted Boltzmann machine**: Only inter-layer pairwise interactions between visible and hidden nodes are assigned a weight

Figure 2.1 Architectures of early energy-based models. The visible nodes (gray) are placeholders for input data, and the hidden nodes (red) are free variables of the model. [1]

By early 2010, the use of GPUs to accelerate deep learning (Raina et al., 2009; Cireşan et al., 2011) made neural networks practical for a variety of complex problems. Several significant deep generative models (generative models based on deep neural networks) were proposed including the variational autoencoder (Kingma et al., 2014b), and generative adversarial networks (Goodfellow et al., 2014). These models were able to leverage the expressive power of neural networks to learn complex real-life distributions.

---

[1] Figure adapted from Patrick Huembeli's tutorial: https://physicsofebm.github.io/

As such, the next step for EBMs was to express the energy function as a deep neural network. However, during this time, research on EBMs remained somewhat stagnant; the issue was that training and inference of these models were inefficient. In particular, approximate inference using Metropolis-Hastings was not scalable to many high-dimensional real-life problems. It was not until several years later when (Du et al., 2019; Song et al., 2019a) utilized gradient-based Langevin dynamics that *deep energy-based models* became more practical. These breakthroughs would further lead to deep EBMs capable of outperforming state-of-the-art GANs at certain image generation tasks (Song et al., 2021a).

Another advantage of deep EBMs is their generality and model flexibility. Unlike previous models, the architecture of the energy function can be selected based on the type of problem. For instance, when generating image data, it is common to use convolutional neural networks since they closely resemble connections in the visual cortex. Generating sequential data in the form of text or audio is more suited to recurrent neural networks since they maintain an internal memory across a temporal space. Similarly, graph neural networks are the natural choice for modelling graph data.

### 2.2.3 EBM training

The biggest challenge for energy-based models is on training and inference. Despite the core EBM training algorithms being discovered decades ago, many of these methods still suffer from the curse of dimensionality. It was only after the emergence of deep learning and GPU programming that some of these methods became practical.

#### 2.2.3.1 Maximum likelihood learning with MCMC

The de-facto method for learning any probabilistic model is with maximum likelihood estimation (MLE). This involves estimating the parameters of the model that are most likely to produce the observed data, and is achieved by maximizing the log-likelihood function over the data distribution with respect to the model parameter space. MLE can also be explained as the minimization of the Kullback-Leibler (KL) divergence between the data and model distribution:

$$
\begin{aligned}
KL(p_{\text{data}} \| p_\theta) &= \mathbb{E}_{p_{\text{data}}(\mathbf{x})} \left[ \log \frac{p_{\text{data}}(\mathbf{x})}{p_\theta(\mathbf{x})} \right] \\
&= - \underbrace{\mathbb{E}_{p_{\text{data}}(\mathbf{x})} \left[ \log p_\theta(\mathbf{x}) \right]}_{\text{MLE objective}} + \text{const.},
\end{aligned}
$$

where the constant is independent of the parameter $\theta$ and can be ignored.

In practice, direct application of MLE is not possible since the model density $p_\theta(\mathbf{x})$ is only known up to a normalizing constant. Alternatively, we can consider the gradients of the maximum likelihood objective:

$$
\begin{aligned}
\nabla_\theta \, \mathbb{E}_{p_{\text{data}}(\mathbf{x})} \left[ \log p_\theta(\mathbf{x}) \right] &= \nabla_\theta \, \mathbb{E}_{p_{\text{data}}(\mathbf{x})} \left[ \log \frac{1}{Z_\theta} e^{-E_\theta(\mathbf{x})} \right] \qquad (2.4) \\
&= -\nabla_\theta \log Z_\theta - \mathbb{E}_{p_{\text{data}}(\mathbf{x})} \left[ \nabla_\theta E_\theta(\mathbf{x}) \right] \\
&= \mathbb{E}_{p_\theta(\mathbf{x})} \left[ \nabla_\theta E_\theta(\mathbf{x}) \right] - \mathbb{E}_{p_{\text{data}}(\mathbf{x})} \left[ \nabla_\theta E_\theta(\mathbf{x}) \right],
\end{aligned}
$$

where the first term on the last line is a result of clever algebraic manipulation of the log-partition term, and is detailed as follows:

$$
\begin{aligned}
\nabla_\theta \log Z_\theta &= \frac{1}{Z_\theta} \int \nabla_\theta e^{-E_\theta(\mathbf{x})} \, \mathrm{d}\mathbf{x} \\
&= \frac{1}{Z_\theta} \int e^{-E_\theta(\mathbf{x})} (-\nabla_\theta E_\theta(\mathbf{x})) \, \mathrm{d}\mathbf{x} \\
&= -\int \nabla_\theta E_\theta(\mathbf{x}) p_\theta(\mathbf{x}) \, \mathrm{d}\mathbf{x} \\
&= -\mathbb{E}_{p_\theta(\mathbf{x})} \left[ \nabla_\theta E_\theta(\mathbf{x}) \right].
\end{aligned}
$$

The energy gradient term in (2.4) can be determined either by automatic differentiation (deep EBMs) or by manual calculation (Boltzmann machines). Therefore, even though direct computation of the maximum likelihood is not possible, we are able to estimate unbiased gradients of the MLE objective, provided that we can sample from $p_{\text{data}}(\mathbf{x})$ and $p_\theta(\mathbf{x})$. These Monte Carlo estimates make model training feasible through gradient-based optimization methods like stochastic gradient descent (SGD).

While sampling from the data distribution $p_{\text{data}}(\mathbf{x})$ is straightforward, sampling from the model distribution $p_\theta(\mathbf{x})$ is challenging. The obvious approach is to use MCMC methods, but not all of these methods scale well to higher dimensions. In the case of deep EBMs on continuous data, we can rely on several gradient based MCMC samplers such as Langevin Monte Carlo and Hamiltonian Monte Carlo. These methods make use of the score function (gradient log-density) to guide the sampler to regions of higher probability.

In Langevin Monte Carlo, we simulate an overdamped Langevin diffusion process with steady state distribution $p_\theta(\mathbf{x})$ and step size $\tau$. Each step $k$ of the simulation is determined with the update:

$$
\mathbf{x}_{k+1} = \mathbf{x}_k + \frac{\tau}{2} \nabla_{\mathbf{x}} \log p_\theta(\mathbf{x}_k) + \sqrt{\tau} \, \boldsymbol{\epsilon} \quad \text{where } \boldsymbol{\epsilon} \sim \mathcal{N}(\mathbf{0}, \mathbf{I}). \qquad (2.5)
$$

Normally, we would need to apply a Metropolis-Hastings accept-reject step to account for the discretization error; however, in practice, if the step size $\epsilon$ is small enough, this error is negligible. As such, when $K \to \infty$ and $\tau \to 0$, the points $\mathbf{x}_K$ converge to samples from the stationary distribution $p_\theta(\mathbf{x})$. This process closely resembles SGD with an added noise component that is proportional to the step size. Also worth noting is that the score function of EBMs is the negative gradient energy (i.e., $\nabla_\theta \log p_\theta(\mathbf{x}) = -\nabla_\theta E_\theta(\mathbf{x})$), making each step straightforward to calculate.

For Boltzmann machines, gradient based samplers are unreliable since the model state space is binary. Instead, we opt to use Gibbs sampling; although not as efficient as gradient based MCMC, Gibbs sampling is a special case of Metropolis-Hastings where the proposal acceptance rate is 1. As such, it is effective for models in which all full-conditional probabilities are known and easy to compute. Fortunately, for Boltzmann machines, the full-conditionals can be calculated analytically as:

$$p(x_i = +1|\mathbf{x}_{-i}) = \sigma\left(\Delta E_i(\mathbf{x})\right) = \sigma\left(2 \sum_j w_{ij} x_j + 2b_i\right),$$

where $\mathbf{x} = (\mathbf{v}, \mathbf{h})$ contains both visible and hidden nodes, $\sigma$ denotes the logistic sigmoid function, and $\Delta E_i = E_{i=\text{off}} - E_{i=\text{on}}$ is the energy difference that results from a single unit $x_i$ being off (-1) versus on (+1). Therefore, we can apply Gibbs sampling by first sampling node $i$ with the given activation probability $p(x_i|\mathbf{x}_{-i})$, while keeping all other nodes fixed. Next, we sample from the subsequent node $i + 1$, while fixing all other nodes (including ones recently updated). This process is repeated an adequate number of times while maintaining the sampling order. Despite being faster than Metropolis-Hastings, Gibbs sampling from Boltzmann machines is still impractical for large models. This is because every step of the sampler requires us to compute a matrix multiplication that scales according to the model size squared $O(D^2)$.

To alleviate this issue, restricted Boltzmann machines make sampling more efficient by only allowing inter-layer connections between nodes. By doing this, the visible nodes become mutually independent given the hidden nodes and vice versa. Since each update of a visible node now does not depend on other visible nodes, the computational complexity of each Gibbs step is significantly reduced. Moreover, the calculation of RBM activation probabilities remains largely the same as with Boltzmann machines, and the only difference is that we now consider full-conditionals for both visible and hidden nodes. Analytic expressions for these conditionals are given below:

$$p(v_i = +1|\mathbf{v}_{-i}, \mathbf{h}) = \sigma\left(\Delta E_i(\mathbf{v}, \mathbf{h})\right) = \sigma\underbrace{\left(2\sum_j w_{ij}h_j + 2a_i\right)}_{\text{indep. of } \mathbf{v}_{-i}}$$

$$p(h_j = +1|\mathbf{v}, \mathbf{h}_{-j}) = \sigma\left(\Delta E_j(\mathbf{v}, \mathbf{h})\right) = \sigma\underbrace{\left(2\sum_i w_{ij}v_i + 2b_j\right)}_{\text{indep. of } \mathbf{h}_{-j}},$$

where $\Delta E_i$ is the resulting energy difference when visible node $v_i$ is off (-1) versus on (+1), and $\Delta E_j$ is the corresponding energy difference for hidden node $h_j$. In fact, this explains why visible nodes are mutually independent given the hidden nodes: the energy difference of any node $i$ is irrespective of the values of other nodes in the layer. Since all activation probabilities are known, we can once again use Gibbs sampling. But this time, rather than updating one node each step, we can now simultaneously update an entire layer. This sampling strategy is outlined below:

1. Fix the visible nodes and sample the hidden nodes from $p(\mathbf{h}|\mathbf{v}) = \Pi_j p(h_j|\mathbf{v})$.

2. Fix the hidden nodes and sample the visible nodes from $p(\mathbf{v}|\mathbf{h}) = \Pi_i p(v_i|\mathbf{h})$.

3. Repeat the above steps until steady-state.

Furthermore, each step now only requires us to compute the multiplication of a $V \times H$ matrix, where $V$ and $H$ are the number of the visible and hidden nodes, respectively. This reduces the complexity per step from $O(D^2)$ to $O(VH)$.

Although MCMC allows us to draw samples from the model distribution, running the sampler to convergence is computationally expensive. For instance, the discrete Langevin diffusion process is shown (under certain assumptions) to produce a distribution $p$ with $KL(p\|p^*) \leq \epsilon$ in $\tilde{O}(\frac{d}{\epsilon})$ steps, where $d$ is the state space dimension (Cheng et al., 2018). This means that the sampler's complexity is inversely proportional to the sampler's accuracy. In practice, we choose to sacrifice convergence for efficiency by using approximate MCMC methods. Contrastive divergence (CD) (Hinton, 2002) is one such method in which MCMC chains are initialized from the data distribution and then run for a fixed number of steps (typically less than required for convergence). Persistent contrastive divergence (Tieleman, 2008) is an improvement on CD in which a single chain is used to initialize all samples across all update steps: each new sample in the training process is used to initialize the chain for subsequent samples. Du et al. (2019) further improve on this by keeping a sample replay buffer from which new chains are randomly sampled.

### 2.2.3.2   Score matching

The score of a continuous distribution $p(\mathbf{x})$ is defined as the gradient of the log-density function $\nabla_{\mathbf{x}} \log p(\mathbf{x})$. Particularly useful about this function is that it is invariant to the normalizing constant, since $\nabla_{\mathbf{x}} \log (Z \cdot p(\mathbf{x})) = \nabla_{\mathbf{x}} \log p(\mathbf{x})$ for any choice of constant $Z$. As such, the score of an EBM does not involve the typically intractable partition function $Z_\theta$, and is an enticing expression to deal with.

If the score functions of two continuously differentiable densities are equal everywhere, then the two densities are also equal everywhere. *Score matching* (SM) (Hyvärinen, 2005) applies this concept by matching the score function of the model to the data, thereby forcing the model distribution to learn the data distribution. The vanilla SM objective, also known as the Fisher divergence, minimizes the expected mean squared error between the model score and data score over the data distribution and is given by:

$$J_{\text{SM}}(\theta) = \mathbb{E}_{p_{\text{data}}(\mathbf{x})} \left[ \frac{1}{2} \| \nabla_{\mathbf{x}} \log p_{\text{data}}(\mathbf{x}) - \nabla_{\mathbf{x}} \log p_\theta(\mathbf{x}) \|^2 \right]. \tag{2.6}$$

Although the expectation can be easily estimated through Monte Carlo simulation, the data score term inside the expectation is unknown and problematic. Fortunately, with clever manipulation, we can derive an alternative objective that only depends on the model distribution:

$$\begin{aligned}
J_{\text{SM}}(\theta) &= \mathbb{E}_{p_{\text{data}}} \left[ \frac{1}{2} \| \nabla_{\mathbf{x}} \log p_{\text{data}}(\mathbf{x}) \|^2 + \frac{1}{2} \| \nabla_{\mathbf{x}} \log p_\theta(\mathbf{x}) \|^2 - \nabla_{\mathbf{x}} \log p_{\text{data}}(\mathbf{x})^\top \nabla_{\mathbf{x}} \log p_\theta(\mathbf{x}) \right] \\
&= \mathbb{E}_{p_{\text{data}}} \left[ \frac{1}{2} \| \nabla_{\mathbf{x}} \log p_\theta(\mathbf{x}) \|^2 - \nabla_{\mathbf{x}} \log p_{\text{data}}(\mathbf{x})^\top \nabla_{\mathbf{x}} \log p_\theta(\mathbf{x}) \right] + \text{const.} \\
&= \mathbb{E}_{p_{\text{data}}} \left[ \frac{1}{2} \| \nabla_{\mathbf{x}} \log p_\theta(\mathbf{x}) \|^2 \right] - \int \nabla_{\mathbf{x}} p_{\text{data}}(\mathbf{x})^\top \nabla_{\mathbf{x}} \log p_\theta(\mathbf{x}) \, d\mathbf{x} + \text{const.},
\end{aligned}$$

where in the second line we absorb all the terms independent of $\theta$ out of the expression. To address the intractable integral of the last line, we can manipulate the product rule for a scalar-vector product:

$$\begin{aligned}
\nabla_{\mathbf{x}} \left( p_{\text{data}}(\mathbf{x}) \cdot \nabla_{\mathbf{x}} \log p_\theta(\mathbf{x}) \right) &= \nabla_{\mathbf{x}} p_{\text{data}}(\mathbf{x}) \cdot \nabla_{\mathbf{x}} \log p_\theta(\mathbf{x})^\top + p_{\text{data}}(\mathbf{x}) \cdot \nabla_{\mathbf{x}}^2 \log p_\theta(\mathbf{x}) \\
p_{\text{data}}(\mathbf{x}) \cdot \nabla_{\mathbf{x}} \log p_\theta(\mathbf{x}) \Big|_{\partial \mathcal{X}} &= \int \left( \nabla_{\mathbf{x}} p_{\text{data}}(\mathbf{x}) \cdot \nabla_{\mathbf{x}} \log p_\theta(\mathbf{x})^\top + p_{\text{data}}(\mathbf{x}) \cdot \nabla_{\mathbf{x}}^2 \log p_\theta(\mathbf{x}) \right) d\mathbf{x} \\
0 &= \int \nabla_{\mathbf{x}} p_{\text{data}}(\mathbf{x}) \cdot \nabla_{\mathbf{x}} \log p_\theta(\mathbf{x})^\top \, d\mathbf{x} \\
&\quad + \int p_{\text{data}}(\mathbf{x}) \cdot \nabla_{\mathbf{x}}^2 \log p_\theta(\mathbf{x}) \, d\mathbf{x}.
\end{aligned}$$

This expression results from taking definite integrals over the entire state space $\mathcal{X}$, for which the left-hand side equates to zero whenever $\lim_{\|\mathbf{x}\| \to \infty} p_{\text{data}}(\mathbf{x}) \nabla_{\mathbf{x}} \log p_\theta(\mathbf{x}) = 0$.

Next, we can take the trace of both sides to get:

$$\int \text{Tr} \left( \nabla_\mathbf{x} p_{\text{data}}(\mathbf{x}) \nabla_\mathbf{x} \log p_\theta(\mathbf{x})^\mathsf{T} \right) d\mathbf{x} = - \int p_{\text{data}}(\mathbf{x}) \cdot \text{Tr} \left( \nabla_\mathbf{x}^2 \log p_\theta(\mathbf{x}) \right) d\mathbf{x}$$

$$\int \text{Tr} \left( \nabla_\mathbf{x} p_{\text{data}}(\mathbf{x})^\mathsf{T} \nabla_\mathbf{x} \log p_\theta(\mathbf{x}) \right) d\mathbf{x} = - \mathbb{E}_{p_{\text{data}}(\mathbf{x})} \left[ \text{Tr} \left( \nabla_\mathbf{x}^2 \log p_\theta(\mathbf{x}) \right) \right]$$

$$\int \nabla_\mathbf{x} p_{\text{data}}(\mathbf{x})^\mathsf{T} \nabla_\mathbf{x} \log p_\theta(\mathbf{x}) \, d\mathbf{x} = - \mathbb{E}_{p_{\text{data}}(\mathbf{x})} \left[ \text{Tr} \left( \nabla_\mathbf{x}^2 \log p_\theta(\mathbf{x}) \right) \right],$$

where the left-hand side of the second line is a result of the cyclic property of trace (i.e., trace trick), and the corresponding term on the last line is because the trace of a scalar is itself. Using this expression, we can now formulate the *implicit score matching* objective:

**Theorem 2.1:** *If the model score function $\nabla_\mathbf{x} \log p_\theta(\mathbf{x})$ is continuously differentiable, then the score matching objective in (2.6) for a differentiable data distribution $p_{data}(\mathbf{x})$ can be expressed as*

$$J_{\text{SM}}(\theta) = \mathbb{E}_{p_{\text{data}}(\mathbf{x})} \left[ \frac{1}{2} \| \nabla_\mathbf{x} \log p_\theta(\mathbf{x}) \|^2 + \text{Tr} \left( \nabla_\mathbf{x}^2 \log p_\theta(\mathbf{x}) \right) \right] + \text{const.}$$

*Estimators of this objective are consistent assuming that the following weak regularity conditions are satisfied:*

- $\mathbb{E}_{p_\theta(\mathbf{x})} [\| \nabla_\mathbf{x} \log p_\theta(\mathbf{x}) \|^2] < \infty \quad \text{and} \quad \mathbb{E}_{p_{\text{data}}(\mathbf{x})} [\| \nabla_\mathbf{x} \log p_{\text{data}}(\mathbf{x}) \|^2] < \infty$
- $\lim_{\|\mathbf{x}\| \to \infty} p_{\text{data}}(\mathbf{x}) \nabla_\mathbf{x} \log p_\theta(\mathbf{x}) = 0.$

Implicit score matching remarkably allows us to obtain unbiased estimators of the Fisher divergence, despite not knowing the explicit data score. Unfortunately, its biggest drawback is the computation of the Hessian term which scales quadratically with the dimension size.

One method for sidestepping this difficulty is by considering a noisy data distribution $p_\sigma(\tilde{\mathbf{x}}) = \int p_\sigma(\tilde{\mathbf{x}}|\mathbf{x}) p_{\text{data}}(\mathbf{x}) \, d\mathbf{x}$, where $p_\sigma(\tilde{\mathbf{x}}|\mathbf{x}) = \mathcal{N}(\tilde{\mathbf{x}}; \mathbf{x}, \sigma^2 \mathbf{I})$ is a Gaussian kernel with noise $\sigma$. This distribution can also be thought of as a non-parametric Parzen density estimate of the data. The natural next step is to consider learning a non-parametric estimate of the data density, and indeed Vincent (2011) shows that score matching between the model distribution $p_\theta(\tilde{\mathbf{x}})$ and the Parzen window estimate of the data $p_\sigma(\tilde{\mathbf{x}})$ yields the following objective:

$$\begin{aligned} J_{\text{DSM}}(\theta) &= \mathbb{E}_{p_\sigma(\tilde{\mathbf{x}})} \left[ \frac{1}{2} \| \nabla_{\tilde{\mathbf{x}}} \log p_\sigma(\tilde{\mathbf{x}}) - \nabla_{\tilde{\mathbf{x}}} \log p_\theta(\tilde{\mathbf{x}}) \|^2 \right] \\ &= \mathbb{E}_{p_\sigma(\tilde{\mathbf{x}}, \mathbf{x})} \left[ \frac{1}{2} \| \nabla_{\tilde{\mathbf{x}}} \log p_\sigma(\tilde{\mathbf{x}}|\mathbf{x}) - \nabla_{\tilde{\mathbf{x}}} \log p_\theta(\tilde{\mathbf{x}}) \|^2 \right] + \text{const.} \end{aligned} \quad (2.7)$$

This objective is called *denoising score matching* (DSM) and completely avoids both the unknown $p_{\text{data}}(\mathbf{x})$ and the computationally expensive $\nabla_{\mathbf{x}}^2 \log p_\theta(\mathbf{x})$ term in implicit SM. The proof is given in the appendix A.1. Regular score matching assumes several regularity conditions on $\log p_{\text{data}}(\mathbf{x})$, including being continuously differentiable and finite everywhere. This makes SM inapplicable to discrete or bounded distributions. On the other hand, DSM addresses this issue by adding a bit of noise to each datapoint $\tilde{\mathbf{x}} = \mathbf{x} + \epsilon$. Since the added noise $\epsilon$ is a continuous Gaussian, the distribution $p_\sigma(\tilde{\mathbf{x}})$ is assured to be both continuously differentiable and non-zero everywhere. This allows DSM to be defined for both discrete and bounded densities. However, it should be noted that DSM does not learn the original data distribution, but rather a noisy version of it; as such, the performance of denoising score matching is very sensitive to the choice of kernel bandwidth $\sigma$.

An alternative way to make implicit score matching scalable is by using stochastic trace estimators. The *Hutchinson trace estimator* (Hutchinson, 1989) is one such example. It is an unbiased Monte Carlo estimator of the trace of any square matrix, and is based on the expectations of quadratic forms $\mathbb{E}[\mathbf{z}\mathbf{z}^{\mathsf{T}}] = \Sigma + \boldsymbol{\mu}\boldsymbol{\mu}^{\mathsf{T}}$, where $\mathbf{z}$ is a random vector with mean $\boldsymbol{\mu}$ and covariance $\Sigma$. If $\mathbf{z}$ has zero-mean unit-variance, then the trace of any square matrix $A$ is given by:

$$\text{Tr}(A) = \text{Tr}(A\,\mathbb{E}[\mathbf{z}\mathbf{z}^{\mathsf{T}}])$$
$$= \mathbb{E}[\text{Tr}(A\mathbf{z}\mathbf{z}^{\mathsf{T}})]$$
$$= \mathbb{E}[\text{Tr}(\mathbf{z}^{\mathsf{T}}A\mathbf{z})] = \mathbb{E}[\mathbf{z}^{\mathsf{T}}A\mathbf{z}],$$

where the first line is multiplied by $\mathbb{E}[\mathbf{z}\mathbf{z}^{\mathsf{T}}] = I$, the second line is due to the linearity of the expectation, and the last line is an application of the trace trick. Applying the Hutchinson estimator for the trace-Hessian term in implicit score matching leads to the following objective:

$$J_{\text{SM}}(\theta) = \mathbb{E}_{p_{\text{data}}(\mathbf{x})} \left[ \frac{1}{2} \| \nabla_{\mathbf{x}} \log p_\theta(\mathbf{x}) \|^2 + \text{Tr}\left( \nabla_{\mathbf{x}}^2 \log p_\theta(\mathbf{x}) \right) \right] \qquad (2.8)$$
$$= \mathbb{E}_{p_{\text{data}}(\mathbf{x})p(\mathbf{v})} \left[ \frac{1}{2} \| \nabla_{\mathbf{x}} \log p_\theta(\mathbf{x}) \|^2 + \mathbf{v}^{\mathsf{T}} \nabla_{\mathbf{x}}^2 \log p_\theta(\mathbf{x})\mathbf{v} \right]$$

where $\mathbf{v}$ is any random vector with zero-mean unit-variance (often Gaussian or Rademacher). At first it seems that computing the Hessian is still unavoidable; however, we can decompose the vector-Hessian product into two gradient operations for which auto-differentiation tools enable efficient calculation: $\mathbf{v}^{\mathsf{T}} \nabla_{\mathbf{x}}^2 \log p_\theta(\mathbf{x}) = \nabla_{\mathbf{x}}\left( \mathbf{v}^{\mathsf{T}} \nabla_{\mathbf{x}} \log p_\theta(\mathbf{x}) \right)$. If $\mathbf{x}$ has dimension $d$, then computing the Hessian on the left would require $d + 1$ gradient operations, whereas the decomposition on the right only requires two gradient operations.

Rather than directly match the scores of the model and data distributions, we can instead match scalar projections of these scores onto random directions. For example, *sliced score matching* (SSM) (Song et al., 2019b) computes the squared error between $\mathbf{v}^\mathsf{T} \nabla_\mathbf{x} \log p_\text{data}(\mathbf{x})$ and $\mathbf{v}^\mathsf{T} \nabla_\mathbf{x} \log p_\theta(\mathbf{x})$ across all random projections $\mathbf{v}$. Similar to implicit score matching (2.1), this objective has an implicit form that does not depend on $p_\text{data}(\mathbf{x})$ and assumes some weak regularity conditions [1]. Many of the steps in this derivation are the same as for implicit score matching and so only the final result is presented below:

$$J_\text{SSM}(\theta) = \mathbb{E}_{p_\text{data}(\mathbf{x})p(\mathbf{v})} \left[ \frac{1}{2} \left( \mathbf{v}^\mathsf{T} \nabla_\mathbf{x} \log p_\text{data}(\mathbf{x}) - \mathbf{v}^\mathsf{T} \nabla_\mathbf{x} \log p_\theta(\mathbf{x}) \right)^2 \right] \quad (2.9)$$

$$= \mathbb{E}_{p_\text{data}(\mathbf{x})p(\mathbf{v})} \left[ \frac{1}{2} \left( \mathbf{v}^\mathsf{T} \nabla_\mathbf{x} \log p_\theta(\mathbf{x}) \right)^2 + \mathbf{v}^\mathsf{T} \nabla_\mathbf{x}^2 \log p_\theta(\mathbf{x})\mathbf{v} \right] + \text{const.}$$

The first line is the explicit sliced score matching objective and the last line is the corresponding implicit form. Unbiased estimators exist for both expressions so long as we can sample from the data $\mathbf{x} \sim p_\text{data}(\mathbf{x})$ and projection $\mathbf{v} \sim p(\mathbf{v})$. Particularly noteworthy is that when $\mathbf{v}$ has zero-mean unit-variance, the implicit SSM objective reduces to implicit score matching with Hutchinson's trace estimator (2.8). This is because $\mathbb{E}_\mathbf{v}[(\mathbf{v}^\mathsf{T} \nabla_\mathbf{x} \log p_\theta(\mathbf{x}))^2] = \|\nabla_\mathbf{x} \log p_\theta(\mathbf{x})\|^2$, the proof of which is given in the appendix A.2. In practice, SSM regularly has higher variance than Hutchinson's estimator; however, its computational efficiency can be further improved by considering finite differences of directional derivatives (Pang et al., 2020). Moreover, several variations of score matching and estimation have been proposed to deal with energy-based latent variable models (Bao et al., 2020, 2021). Models trained with score matching have found recent success at generating high-quality images (Song et al., 2019a, 2021b). This success is largely due to the training simplicity and effectiveness of score matching. Unlike maximum likelihood learning, we do not need to sample from the model during training; and unlike noise-contrastive estimation, we do not need to select a noise distribution.

### 2.2.3.3 Noise-contrastive estimation

One of the most subconscious ways that intelligent beings learn is through comparing and contrasting. *Noise-contrastive estimation* (NCE) (Gutmann and Hyvärinen, 2010) is an application of this principle to generative modelling. The basic idea is to learn a model distribution by contrasting it with some known noise distribution; in particular, we estimate the model parameters by learning (through logistic regression) to discriminate between the observed data and some artificially generated noise.

---

[1] The same regularity conditions as for implicit score matching plus the assumption that $\mathbb{E}[\mathbf{v}\mathbf{v}^\mathsf{T}]$ is positive definite.

The most common form of NCE is based on binary logistic regression. Let us consider two mixture distributions: one between noise and data $p_{\text{n,data}}(\mathbf{x}) = p_{\text{n}}(\mathbf{x})P(\pi = 0) + p_{\text{data}}(\mathbf{x})P(\pi = 1)$, and one between noise and model $p_{\text{n},\theta}(\mathbf{x}) = p_{\text{n}}(\mathbf{x})P(\pi = 0) + p_\theta(\mathbf{x})P(\pi = 1)$. For both mixtures, $\pi$ is a Bernoulli random variable and $p_{\text{n}}(\mathbf{x})$ is some known distribution from which we can easily sample. The posterior probability of $\pi = 0$ (i.e., the probability that a sample $\mathbf{x}$ originated from the noise distribution $p_{\text{n}}(\mathbf{x})$) for both mixture distributions is given by Bayes' theorem as:

$$p_{\text{n,data}}(\pi = 0|\mathbf{x}) = \frac{p_{\text{n,data}}(\mathbf{x}|\pi = 0)P(\pi = 0)}{p_{\text{n,data}}(\mathbf{x}|\pi = 0)P(\pi = 0) + p_{\text{n,data}}(\mathbf{x}|\pi = 1)P(\pi = 1)} \tag{2.10}$$

$$= \frac{p_{\text{n}}(\mathbf{x})}{p_{\text{n}}(\mathbf{x}) + \upsilon p_{\text{data}}(\mathbf{x})},$$

$$p_{\text{n},\theta}(\pi = 0|\mathbf{x}) = \frac{p_{\text{n},\theta}(\mathbf{x}|\pi = 0)P(\pi = 0)}{p_{\text{n},\theta}(\mathbf{x}|\pi = 0)P(\pi = 0) + p_{\text{n},\theta}(\mathbf{x}|\pi = 1)P(\pi = 1)} \tag{2.11}$$

$$= \frac{p_{\text{n}}(\mathbf{x})}{p_{\text{n}}(\mathbf{x}) + \upsilon p_\theta(\mathbf{x})},$$

where $\upsilon = \frac{P(\pi=1)}{P(\pi=0)}$ is the ratio of observed samples to noisy samples. Gutmann and Hyvärinen (2010) show that by fitting the posterior model mixture $p_{\text{n},\theta}(\pi|\mathbf{x})$ to the posterior data mixture $p_{\text{n,data}}(\pi|\mathbf{x})$, the model will indirectly learn the data distribution. We can informally show this to be true by equating the two probabilities:

$$p_{\text{n,data}}(\pi = 0|\mathbf{x}) \equiv p_{\text{n},\theta^*}(\pi = 0|\mathbf{x})$$

$$\implies \frac{p_{\text{n}}(\mathbf{x})}{p_{\text{n}}(\mathbf{x}) + \upsilon p_{\text{data}}(\mathbf{x})} \equiv \frac{p_{\text{n}}(\mathbf{x})}{p_{\text{n}}(\mathbf{x}) + \upsilon p_{\theta^*}(\mathbf{x})}$$

$$\implies p_{\text{data}}(\mathbf{x}) \equiv p_{\theta^*}(\mathbf{x}).$$

Therefore, it suffices to learn the data distribution through standard KL divergence minimization between $p_{\text{n,data}}(\pi|\mathbf{x})$ and $p_{\text{n},\theta}(\pi|\mathbf{x})$, essentially performing binary logistic regression to model the probability that $\mathbf{x}$ comes from observed data rather than noise.

$$J_{\text{NCE}}(\theta) = \mathbb{E}_{p_{\text{n,data}}(\mathbf{x})}\left[KL(p_{\text{n,data}}(\pi|\mathbf{x})\|p_{\text{n},\theta}(\pi|\mathbf{x}))\right] \tag{2.12}$$

$$= -\mathbb{E}_{p_{\text{n,data}}(\mathbf{x},\pi)}\left[\log p_{\text{n},\theta}(\pi|\mathbf{x})\right] + \text{const.}$$

Given the joint samples $(\mathbf{x}_i, \pi_i)_{1..N}$ of observed data and their labels, the estimation of this objective is equivalent to calculating the binary cross-entropy loss:

$$J_{\text{NCE}}(\theta) \simeq -\frac{1}{N} \sum_{i=1}^{N} [\![\pi_i = 0]\!] \log p_{\text{n},\theta}(\pi = 0 | \mathbf{x}_i) + [\![\pi_i = 1]\!] \log p_{\text{n},\theta}(\pi = 1 | \mathbf{x}_i),$$

where the posterior probabilities are given in (2.11). Although this estimator is unbiased, we can further reduce its variance by using the fact the class probabilities $\rho_i := P(\pi = i)$ are known:

$$\begin{aligned} J_{\text{NCE}}(\theta) &= -\mathbb{E}_{p(\pi)} \left[ \mathbb{E}_{p_{\text{n,data}}(\mathbf{x}|\pi)} \left[ \log p_{\text{n},\theta}(\pi | \mathbf{x}) \right] \right] & (2.13) \\ &= -\rho_0 \mathbb{E}_{p_{\text{n}}(\mathbf{x})} \left[ \log p_{\text{n},\theta}(\pi = 0 | \mathbf{x}) \right] - \rho_1 \mathbb{E}_{p_{\text{data}}(\mathbf{x})} \left[ \log p_{\text{n},\theta}(\pi = 1 | \mathbf{x}) \right] \\ &= -\rho_0 \mathbb{E}_{p_{\text{n}}(\mathbf{x})} \left[ \log \frac{p_{\text{n}}(\mathbf{x})}{p_{\text{n}}(\mathbf{x}) + \upsilon p_\theta(\mathbf{x})} \right] - \rho_1 \mathbb{E}_{p_{\text{data}}(\mathbf{x})} \left[ \log \frac{\upsilon p_\theta(\mathbf{x})}{p_{\text{n}}(\mathbf{x}) + \upsilon p_\theta(\mathbf{x})} \right] \end{aligned}$$

where $\upsilon$ is the ratio of the observed samples to noisy samples. A particularly powerful feature of NCE is that the model $p_\theta(\mathbf{x})$ needs not to be normalized. In fact, Gutmann and Hyvärinen (2010)[Theorem 1] show that the optimal model trained using NCE will be *self-normalizing*. Furthermore, it is also possible to explicitly set a learnable partition function $Z_\theta$, such that when the model is trained with NCE, $Z_\theta$ will guarantee that the model integrates to one.

Another variation of noise-contrastive estimation is based on multinomial logistic regression (aka., softmax regression). Rather than learning a binary classifier for observed and noisy data, we instead learn a multinomial classifier for discriminating a single observed sample from many noisy samples. This form of NCE is also referred to as *rank-based noise-contrastive estimation* (Jozefowicz et al., 2016), since softmax regression provides rankings of posterior probabilities. In rank-based NCE, we consider a collection $X = \{\mathbf{x}_1, ..., \mathbf{x}_L\}$ of $L - 1$ noisy samples and a single observed sample. The categorical random variable $\pi \sim \text{Cat}(\rho_1, ..., \rho_L)$ indicates which of the $L$ elements is the observed sample. The noise-data and noise-model mixture distributions that describe this collection are given by:

$$p_{\text{n,data}}(X) = \sum_{i=1}^{L} P(\pi = i) \left( p_{\text{data}}(\mathbf{x}_i) \prod_{\ell \neq i} p_{\text{n}}(\mathbf{x}_l) \right)$$

$$p_{\text{n},\theta}(X) = \sum_{i=1}^{L} P(\pi = i) \left( p_\theta(\mathbf{x}_i) \prod_{\ell \neq i} p_{\text{n}}(\mathbf{x}_l) \right).$$

The posterior probability that element $\mathbf{x}_k$ is the observed sample is determined using

Bayes' theorem to be:

$$p_{\text{n,data}}(\pi = k | X) = \frac{p_{\text{n,data}}(X | \pi = k) P(\pi = k)}{\sum_j p_{\text{n,data}}(X | \pi = j) P(\pi = j)} \tag{2.14}$$

$$= \frac{\rho_k p_{\text{data}}(\mathbf{x}_k) \prod_{\ell \neq k} p_{\text{n}}(\mathbf{x}_\ell)}{\sum_j \rho_j p_{\text{data}}(\mathbf{x}_j) \prod_{\ell \neq j} p_{\text{n}}(\mathbf{x}_\ell)} = \frac{\rho_k \frac{p_{\text{data}}(\mathbf{x}_k)}{p_{\text{n}}(\mathbf{x}_k)}}{\sum_j \rho_j \frac{p_{\text{data}}(\mathbf{x}_j)}{p_{\text{n}}(\mathbf{x}_j)}}$$

$$= \sigma \left( \left[ \rho_i \frac{p_{\text{data}}(\mathbf{x}_i)}{p_{\text{n}}(\mathbf{x}_i)} \right]_{i=1..L} \right)_k$$

$$p_{\text{n},\theta}(\pi = k | X) = \sigma \left( \left[ \rho_i \frac{p_\theta(\mathbf{x}_i)}{p_{\text{n}}(\mathbf{x}_i)} \right]_{i=1..L} \right)_k \tag{2.15}$$

where $\sigma(\cdot)_k$ denotes the softmax function at index $k$, and the equivalence on the second line is a result of multiplying both numerator and denominator by $\prod p_{\text{n}}(\mathbf{x}_\ell)$. Similar to binary NCE, fitting the posterior model mixture $p_{\text{n},\theta}(\pi | X)$ to the posterior data mixture $p_{\text{n,data}}(\pi | X)$ will indirectly fit $p_\theta(\mathbf{x})$ to $p_{\text{data}}(\mathbf{x})$. As such, we once again optimize the maximum likelihood objective (2.12). For a dataset of $N$ collections $X = \{X^{(1)}, ..., X^{(N)}\}$ and their corresponding labels $\{\pi_1, ..., \pi_N\}$, this turns out to be equivalent to calculating the cross-entropy loss:

$$J_{\text{NCE}}(\theta) = - \mathbb{E}_{p_{\text{n,data}}(X, \pi)} \left[ \log p_{\text{n},\theta}(\pi | X) \right]$$

$$\simeq - \sum_{i=1}^{N} \sum_{k=1}^{L} [\![ \pi_i = k ]\!] \log p_{\text{n},\theta}(\pi_i = k | X^{(i)}).$$

In addition, just like with binary NCE, if the class probabilities $\rho_k := P(\pi = k)$ are known, we can develop a lower variance estimator by manually calculating the expectation over $\pi$:

$$J_{\text{NCE}}(\theta) = - \mathbb{E}_{p(\pi)} \left[ \mathbb{E}_{p_{\text{n,data}}(X | \pi)} \left[ \log p_{\text{n},\theta}(\pi | X) \right] \right] \tag{2.16}$$

$$= - \sum_{k=1}^{L} \rho_k \, \mathbb{E}_{p_{\text{n,data}}(X | \pi = k)} \left[ \log p_{\text{n},\theta}(\pi = k | X) \right]$$

$$= - \sum_{k=1}^{L} \rho_k \, \mathbb{E}_{p_{\text{data}}(\mathbf{x}_k) \prod_{\ell \neq k} p_{\text{n}}(\mathbf{x}_\ell)} \left[ \log p_{\text{n},\theta}(\pi = k | X) \right],$$

where the posterior probabilities are given in (2.14). In general, (Ma et al., 2018) find that both binary and rank-based NCE perform similarly for a variety of language modelling tasks. However, the rank-based variant is consistent for a broader class of models than the binary variant.

# CHAPTER 3   SELF-ADAPTING NCE

NCE is a powerful tool for turning unsupervised learning problems into supervised ones. Unfortunately, its performance is heavily influenced by the choice of the noise distribution. If the noise distribution $p_n(\mathbf{x})$ differs too much from the data distribution $p_{\text{data}}(\mathbf{x})$, then the classification task is trivial and training is ineffective. As such, NCE performs best when noise samples are difficult to distinguish from data samples, i.e., $p_n(\mathbf{x}) \approx p_{\text{data}}(\mathbf{x})$. This is especially true for structured and high-dimensional data, where simple noise distributions like Gaussians are insufficient for noise-contrastive learning.

In practice, the data distribution is often unknown, and so NCE has struggled to scale to high dimensions. Several methods have attempted to adapt the noise distribution during training, thereby dynamically improving the training performance. This thesis follows the same line of thought. First, we review some recent methods that tune the noise distribution. Next, we detail the main contribution of this thesis, a noise-contrastive estimation algorithm that utilizes a self-adapting $p_n(\mathbf{x})$. Finally, we discuss the advantages and disadvantages of this algorithm compared to previous methods.

## 3.1   Related Work

One way the noise distribution can be tuned close to the data distribution is by allowing the noise to be dependent on the data. *Conditional noise-contrastive estimation* (CNCE) (Ceylan et al., 2018) accomplishes this by generating noisy samples $\mathbf{y}$ from a conditional noise distribution $p_n(\mathbf{y}|\mathbf{x})$, where $\mathbf{x}$ represents samples from the data. It turns out that, analogous to NCE, solving a classification problem involving the noise and data would indirectly fit $p_\theta(\mathbf{x})$ to $p_{\text{data}}(\mathbf{x})$.

Let us define $\mathcal{X}$ to be the data sample space, and $\mathcal{Y}$ to be the noise sample space generated according to $\int p_n(\mathbf{y}|\mathbf{x})p_{\text{data}}(\mathbf{x})\,d\mathbf{x}$. We denote the ordered tuple of data and noise sample by $\mathbf{u} = (\mathbf{u}_1, \mathbf{u}_2)$, where the order of the data sample in the tuple indicates its class assignment $\pi$. For example, if $\mathbf{u}_1 \in \mathcal{X}$ and $\mathbf{u}_2 \in \mathcal{Y}$ then $\mathbf{u}$ has class assignment $\pi = 0$, and if $\mathbf{u}_1 \in \mathcal{Y}$ and $\mathbf{u}_2 \in \mathcal{X}$ then $\mathbf{u}$ has class assignment $\pi = 1$. Consequently, if a tuple $\mathbf{u}$ has ordering $\pi = 0$, then the tuple's probability is given by $p_{n,\text{data}}(\mathbf{u}|\pi = 0) = p_{\text{data}}(\mathbf{u}_1)p_n(\mathbf{u}_2|\mathbf{u}_1)$. The posterior class probability is determined by Bayes' rule to be:

$$p_{\text{n,data}}(\pi = 0|\mathbf{u}) = \frac{p(\pi = 0)p_{\text{n,data}}(\mathbf{u}|\pi = 0)}{p(\pi = 0)p_{\text{n,data}}(\mathbf{u}|\pi = 0) + p(\pi = 1)p_{\text{n,data}}(\mathbf{u}|\pi = 1)} \tag{3.1}$$
$$= \frac{1}{1 + \upsilon \frac{p_{\text{data}}(\mathbf{u}_2)p_n(\mathbf{u}_1|\mathbf{u}_2)}{p_{\text{data}}(\mathbf{u}_1)p_n(\mathbf{u}_2|\mathbf{u}_1)}},$$

where $\upsilon = \frac{p(\pi = 1)}{p(\pi = 0)}$. If $\mathcal{X}$ is instead defined as the model sample space, then we can replace $p_{\text{data}}(\mathbf{x})$ with $p_\theta(\mathbf{x})$ to get the corresponding posterior class probability:

$$p_{\text{n},\theta}(\pi = 0|\mathbf{u}) = \frac{1}{1 + \upsilon \frac{p_\theta(\mathbf{u}_2)p_n(\mathbf{u}_1|\mathbf{u}_2)}{p_\theta(\mathbf{u}_1)p_n(\mathbf{u}_2|\mathbf{u}_1)}}. \tag{3.2}$$

An interesting point about this posterior is that the model partition functions in the denominator cancel; therefore, evaluation of this probability does not require a normalized model. Ceylan et al. (2018) shows that when we fit $p_{\text{n},\theta}(\pi|\mathbf{u})$ to $p_{\text{n,data}}(\pi|\mathbf{u})$ the model will indirectly learn the data distribution up to a normalizing constant. This is in contrast to NCE where the partition function is a learned parameter. Nonetheless, the (unnormalized) model can still be learned by minimizing the negative log-likelihood given in (2.12).

If we let the noise conditional be a Gaussian $\mathcal{N}(\mathbf{y}|\mathbf{x}, \sigma^2 I)$ centered on the data sample, then $p_{\text{n,data}}(\mathbf{y}) = \int p_n(\mathbf{y}|\mathbf{x})p_{\text{data}}(\mathbf{x})\,d\mathbf{x}$ is the Parzen density estimate of the data distribution. Depending on the choice of bandwidth $\sigma$, the noise samples can be made arbitrarily close to the data samples. Unfortunately, in practice, when $\sigma$ is too large or too small the gradients of the objective vanish, and CNCE fails to learn anything meaningful. This problem is compounded by the fact that high-dimensional real-world data distributions often lie on lower-dimensional manifolds, and therefore have very small variance. As such, if we tune the noise distribution to be close to the data distribution (using small bandwidth $\sigma$), then the training gradients often vanish.

*Adversarial contrastive estimation* (ACE) (Bose et al., 2018) is another approach in making the noise distribution adaptive by modeling it with an implicit variational mixture $p_\varphi(\mathbf{x}) = \int p_\varphi(\mathbf{x}|\mathbf{z})p(\mathbf{z})\,d\mathbf{z}$ (a.k.a., noisy neural sampler), where $p_\varphi(\mathbf{x}|\mathbf{z})$ is a variational conditional model and $p(\mathbf{z})$ is some implicit prior. If we let the noise $p_n(\mathbf{x}) = \lambda p_{n_0}(\mathbf{x}) + (1 - \lambda)p_\varphi(\mathbf{x})$ be a mixture of a base noise distribution $p_{n_0}(\mathbf{x})$ and the implicit $p_\varphi(\mathbf{x})$, then the corresponding noise-contrastive objective is:

$$J_{\text{ACE}}(\theta, \varphi) = -\mathbb{E}_{\text{n,data}}(\mathbf{x}, \pi)\left[\log p_{\text{n},\theta}(\pi|\mathbf{x})\right] \tag{3.3}$$
$$= -p(\pi = 0)\,\mathbb{E}_{p_n(\mathbf{x})}\left[h_\theta(\mathbf{x})\right] - p(\pi = 1)\,\mathbb{E}_{p_{\text{data}}(\mathbf{x})}\left[1 - h_\theta(\mathbf{x})\right]$$
$$= -\rho_0\lambda\,\mathbb{E}_{p_{n_0}(\mathbf{x})}\left[h_\theta(\mathbf{x})\right] - \rho_0(1 - \lambda)\,\mathbb{E}_{p_\varphi(\mathbf{x})}\left[h_\theta(\mathbf{x})\right] - \rho_1\,\mathbb{E}_{p_{\text{data}}(\mathbf{x})}\left[1 - h_\theta(\mathbf{x})\right],$$

where $h_\theta(\mathbf{x}) = \log p_{\mathrm{n},\theta}(\pi = 0|\mathbf{x})$ is the posterior class probability given in (2.11), and $\rho_k := p(\pi = k)$ denotes the class probability. Both the model $h_\theta(\mathbf{x})$ and variational sampler $p_\varphi(\mathbf{x})$ can be concurrently trained in a manner similar to generative adversarial networks: we play an adversarial game in which the generator $p_\varphi(\mathbf{x})$ learns to produce samples from $p_{\mathrm{data}}(\mathbf{x})$, and the discriminator $h_\theta(\mathbf{x})$ determines if samples are real (i.e., come from the data distribution) or fake. This game corresponds to the following min-max optimization:

$$\min_\theta \max_\varphi J_{\mathrm{ACE}}(\theta, \varphi).$$

An important thing to note about optimizing objective (3.3) is that it requires us to back-propagate through samples from $p_\varphi(\mathbf{x})$. If the variational conditional model $p_\varphi(\mathbf{x}|\mathbf{z})$ is reparameterizable, this can be typically done with pathwise gradient estimators (a.k.a., reparameterization trick). For example, if we have a Gaussian model $p_\varphi(\mathbf{x}|\mathbf{z}) = \mathcal{N}(\mathbf{x}|\mu_\varphi(\mathbf{z}), \mathrm{diag}(\sigma_\varphi^2(\mathbf{z})))$, then $\mathbf{x}$ can be sampled according to the following deterministic function: $\mathbf{x} = \mu_\varphi(\mathbf{z}) + \sigma_\varphi(\mathbf{z}) \odot \epsilon$, where $\epsilon$ is sampled from a standard Gaussian. In the case that $p_\varphi(\mathbf{x}|\mathbf{z})$ is not reparameterizable, we can instead rely on the more general REINFORCE gradient estimator (Williams, 1992); however, this estimator is known to suffer from high variance.

Unfortunately, evaluation of the discriminator function $h_\theta(\mathbf{x}) = \frac{p_{\mathrm{n}}(\mathbf{x})}{p_{\mathrm{n}}(\mathbf{x}) + \nu p_\theta(\mathbf{x})}$ in objective (3.3) does not have an analytic form because $p_{\mathrm{n}}(\mathbf{x})$ is implicit. As such, we must resort to biased estimators of the objective $J_{\mathrm{ACE}}$ which scale poorly to high-dimensional data. Furthermore, because adversarial training is notoriously tricky, so too is ACE training. For example, mode collapse (a common issue in GANs) also occurs in ACE when the generator $p_\varphi(\mathbf{x})$ produces samples from only a subset of the modes of data. Consequently, the architectures of $p_\varphi$ and $h_\theta$ must be carefully balanced, so one does not overpower the other. We can mitigate mode collapse with techniques such as entropy regularization; however, min-max training stability remains a challenging area of research.

Rather than modelling $p_{\mathrm{n}}$ with a noisy neural sampler, *flow contrastive estimation* (FCE) (Gao et al., 2020) instead models $p_{\mathrm{n}}(\mathbf{x})$ with a flow-based model. In doing so, FCE sacrifices the expressive power of an implicit $p_{\mathrm{n}}(\mathbf{x})$ for the training stability associated with flow models. Because many recent flow models still compare favorably with GANs (Kingma et al., 2018; Ho et al., 2019), this trade-off in expressivity is largely inconsequential. Flow-based models consider a sequence of invertible transformations $f_\phi = f_{\phi_m} \circ f_{\phi_{m-1}} \circ \cdots \circ f_{\phi_1}$, where $\mathbf{x} = f_\phi(\mathbf{z})$ is the cumulative transform of some known noise distribution $\mathbf{z} \sim p_0(\mathbf{z})$. If each of the intermediate transforms are denoted by ran-dom variable $\mathbf{h}_j = f_{\phi_j}(\mathbf{h}_{j-1})$ with $\mathbf{h}_m = \mathbf{x}$ and $\mathbf{h}_0 = \mathbf{z}$, then under the Jacobian change of variables the flow model distribution can be expressed as:

$$p_\phi(\mathbf{x}) = p_0(f_\phi^{-1}(\mathbf{x})) \left| \det \frac{\partial f_\phi^{-1}(\mathbf{x})}{\partial \mathbf{x}} \right|$$

$$= p_0(f_\phi^{-1}(\mathbf{x})) \prod_{j=1}^m \left| \det \frac{\partial \mathbf{h}_{j-1}}{\partial \mathbf{h}_j} \right|.$$

It turns out that we can carefully design the transformations $f_{\phi_j}$ such that the determinant of the Jacobian is simple to compute. In particular, if we choose a transformation whose Jacobian is a triangular matrix, then the determinant of such a Jacobian is simply the product of the diagonal entries. Consequently, flow models allow for tractable likelihoods and stable training.

Just like adversarial contrastive estimation, flow contrastive estimation also employs an adversarial training procedure. The objective we consider is analogous to the original NCE objective (2.13), only this time the noise distribution $p_n(\mathbf{x})$ is replaced by a trainable flow model $p_\phi(\mathbf{x})$:

$$
\begin{aligned}
J_{\text{FCE}}(\theta, \phi) &= -\mathbb{E}_{p_{\phi,\text{data}}(\mathbf{x},\pi)} \left[ \log p_{\phi,\theta}(\pi | \mathbf{x}) \right] \qquad (3.4)\\
&= -\rho_0 \, \mathbb{E}_{p_\phi(\mathbf{x})} \left[ \log p_{\phi,\theta}(\pi = 0 | \mathbf{x}) \right] - \rho_1 \, \mathbb{E}_{p_{\text{data}}(\mathbf{x})} \left[ \log p_{\phi,\theta}(\pi = 1 | \mathbf{x}) \right]\\
&= -\rho_0 \, \mathbb{E}_{p_\phi(\mathbf{x})} \left[ \log \frac{p_\phi(\mathbf{x})}{p_\phi(\mathbf{x}) + \upsilon p_\theta(\mathbf{x})} \right] - \rho_1 \, \mathbb{E}_{p_{\text{data}}(\mathbf{x})} \left[ \log \frac{p_\theta(\mathbf{x})}{p_\theta(\mathbf{x}) + \upsilon^{-1} p_\phi(\mathbf{x})} \right]\\
&= -\rho_0 \, \mathbb{E}_{p_0(\mathbf{z})} \left[ \log \frac{p_\phi(f_\phi(\mathbf{z}))}{p_\phi(f_\phi(\mathbf{z})) + \upsilon p_\theta(f_\phi(\mathbf{z}))} \right] - \rho_1 \, \mathbb{E}_{p_{\text{data}}(\mathbf{x})} \left[ \log \frac{p_\theta(\mathbf{x})}{p_\theta(\mathbf{x}) + \upsilon^{-1} p_\phi(\mathbf{x})} \right],
\end{aligned}
$$

where $\rho_k := p(\pi = k)$. The last line is a result of the law of the unconscious statistician, such that gradients can pass through the expectation via the reparameterization trick. This adversarial objective is then optimized through the following min-max training procedure: $\min_\theta \max_\phi J_{\text{FCE}}(\theta, \phi)$. Because of the tractability of flow-based models, FCE performs better than ACE for many high-dimensional problems; however, the adversarial training procedure still poses challenges and requires careful architecture and hyperparameter tuning.

## 3.2  Method: AdaNCE

Although there exist several ways to tune the noise distribution to be close to the data distribution, they all suffer from several drawbacks. For example, in conditional noise-contrastive estimation, the choice of kernel bandwidth $\sigma$ for noisy samples often leads to vanishing training gradients which renders learning ineffective. Furthermore, in both adversarial and flow contrastive estimation we model the noise distribution with a separate generative model. This noise model is concurrently trained with an adversarial objective, which is known at times to be notoriously difficult to optimize.

|  | CNCE | ACE | FCE | AdaNCE |
|---|---|---|---|---|
| Self-Normalizing | No | Yes | Yes | No |
| Adversarial Training | No | Yes | Yes | No |
| Scalable | No | No | Yes | Yes |

Table 3.1   comparison of the limitations of several noise-adaptive NCE algorithms: self-normalizing indicates if the partition function is learned; adversarial training is whether the algorithm uses an adversarial learning procedure; scalable is whether or not the method is scalable to higher dimensions.

This thesis attempts to make noise-contrastive estimation scalable to higher dimensions while avoiding the difficult-to-optimize adversarial objective. Following the previous line of work, we opt to have an adaptive noise distribution $p_n(\mathbf{x})$ where $p_n(\mathbf{x})$ progressively tunes itself to be close to the data distribution $p_{data}(\mathbf{x})$. However, unlike in adversarial and flow contrastive estimation, we instead propose to use static instances of the EBM learned at previous training iterations as the noise model. The idea is that these static instances progressively converge to the data distribution, since each instance is a better approximation of $p_{data}(\mathbf{x})$ than the previous instance. The proposed algorithm is called *self-adapting noise-contrastive estimation* (AdaNCE). AdaNCE circumvents the need to train $p_n(\mathbf{x})$, and is scalable.

Let us consider a general energy-based model parameterized by some neural network $\theta$. The corresponding Boltzmann distribution that describes this EBM is:

$$p_\theta(\mathbf{x}) = \frac{1}{Z_\theta} e^{-E_\theta(\mathbf{x})},$$

where $Z_\theta = \int e^{-E_\theta(\mathbf{x})} \, d\mathbf{x}$ is the intractable partition function. In AdaNCE, we let the noise distribution $p_m(\mathbf{x}) := p_{\theta_t}(\mathbf{x})$ be the fixed model distribution at some previous iteration $t$. We emphasize that this noise distribution has its parameters frozen so that $p_m(\mathbf{x})$ does not get updated. Just like in noise-contrastive estimation, the objective we minimize is the negative log likelihood of the class posterior. Assuming there are an equal number of noise and data samples, i.e., $p(\pi = 0) = p(\pi = 1) = 0.5$, this objective can be expressed as:

$$
\begin{aligned}
J(\theta) &= -\mathbb{E}_{p_{\mathrm{m,data}}(\mathbf{x},\pi)} \left[ \log p_{\mathrm{m},\theta}(\pi|\mathbf{x}) \right] \qquad\qquad\qquad\qquad (3.5) \\
&= -\frac{1}{2} \mathbb{E}_{p_{\mathrm{m}}(\mathbf{x})} \left[ \log p_{\mathrm{m},\theta}(\pi = 0|\mathbf{x}) \right] - \frac{1}{2} \mathbb{E}_{p_{\mathrm{data}}(\mathbf{x})} \left[ \log p_{\mathrm{m},\theta}(\pi = 1|\mathbf{x}) \right] \\
&\propto -\mathbb{E}_{p_{\mathrm{m}}(\mathbf{x})} \left[ \log \frac{p_{\mathrm{m}}(\mathbf{x})}{p_{\mathrm{m}}(\mathbf{x}) + p_{\theta}(\mathbf{x})} \right] - \mathbb{E}_{p_{\mathrm{data}}(\mathbf{x})} \left[ \log \frac{p_{\theta}(\mathbf{x})}{p_{\mathrm{m}}(\mathbf{x}) + p_{\theta}(\mathbf{x})} \right],
\end{aligned}
$$

where it is possible to obtain unbiased estimators of $J(\theta)$ so long as we can draw samples from $p_{\mathrm{m}}(\mathbf{x})$ and $p_{\mathrm{data}}(\mathbf{x})$. The model parameters $\theta$ can then be learned through minimization of $\hat{J}(\theta)$ with gradient-based optimization methods (e.g., stochastic gradient descent or Adam (Kingma et al., 2014a)). During this training process, we set the noise distribution $p_{\mathrm{m}}(\mathbf{x})$ to be the current (frozen) model distribution $p_{\theta}(\mathbf{x})$ after a specified number of update steps, referred to as the *adaptive interval* $\mathcal{K}$. The basic AdaNCE procedure is summarized in Algorithm 3.1.

---

**Algorithm 3.1** Self-Adapting Noise-Contrastive Estimation

---

**Input:** $E_{\theta}(\cdot)$ energy function, $\mathcal{K}$ adaptive interval, $T$ training iterations, $N$ batch size

**Initialize:** $p_{\mathrm{m}} \leftarrow \mathrm{Freeze}(p_{\theta_0})$; $p_{\theta_0}(\cdot) = \frac{1}{Z_{\theta_0}} e^{-E_{\theta_0}(\cdot)}$ is the initial noise distribution at $t = 0$

1: **for** $t = 0 : T$ **do**
2:      $\mathbf{x}^{+}_{i=1..N} \sim p_{\mathrm{data}}(\mathbf{x})$
3:      $\mathbf{x}^{-}_{i=1..N} \sim p_{\mathrm{m}}(\mathbf{x})$
4:      $\hat{J}(\theta_t) = -\frac{1}{N} \sum_i \left( \log \frac{p_{\mathrm{m}}(\mathbf{x}^{-}_i)}{p_{\mathrm{m}}(\mathbf{x}^{-}_i) + p_{\theta_t}(\mathbf{x}^{-}_i))} + \log \frac{p_{\theta_t}(\mathbf{x}^{+}_i)}{p_{\mathrm{m}}(\mathbf{x}^{+}_i) + p_{\theta_t}(\mathbf{x}^{+}_i)} \right)$
5:      $\theta_{t+1} \leftarrow \theta_t - \eta \nabla \hat{J}(\theta_t)$
6:      **if** $t + 1 \% \mathcal{K} == 0$ **then**
7:          $p_{\mathrm{m}} \leftarrow \mathrm{Freeze}(p_{\theta_{t+1}})$
8:      **end if**
9: **end for**

---

*Note: The assignment $p_{\mathrm{m}} \leftarrow \mathrm{Freeze}(p_{\theta})$ is accomplished by setting the energy function $E_{\theta}(\cdot)$ (with parameters frozen) as the energy function of the noise distribution. In this way, we avoid estimating the intractable partition function.

At first glance, there are two major difficulties with evaluating objective (3.5). The first issue is that we typically only know the noise distribution up to some normalization constant. This is because the partition function of the model distribution $p_{\theta}(\mathbf{x})$ and consequently the partition function of the noise distribution $p_{\mathrm{m}}(\mathbf{x})$ are intractable. Fortunately, it turns out that if we use values of the unnormalized noise density $\tilde{p}_{\mathrm{m}}(\mathbf{x})$ in the AdaNCE objective, the optimal model distribution is still given by the data distribution, i.e., $p_{\theta^*}(\mathbf{x}) = p_{\text{data}}(\mathbf{x})$. However, unlike in noise-contrastive estimation, the model's partition function $Z_\theta$ is not learned, and so $e^{-E_\theta(\mathbf{x})}$ is not guaranteed to be self-normalizing. This property of AdaNCE is further explained in the following chapter 4.3 through the framework of Bregman divergences.

The second challenge in evaluation of objective (3.5) is sampling from the noise distribution. When $p_{\text{m}}(\mathbf{x})$ is a static distribution that is easy to sample from (i.e., has a tractable and invertible cdf), we can use inversion sampling to efficiently draw $\mathbf{x} \sim p_{\text{m}}(\mathbf{x})$. Unfortunately, in our case, the noise $p_{\text{m}}(\mathbf{x}) := \frac{1}{Z_{\theta_t}} e^{-E_{\theta_t}(\mathbf{x})}$ is not only difficult to sample from, but it also adapts and changes. As such, we must rely on the general class of Markov chain Monte Carlo methods to sample from $p_{\text{m}}(\mathbf{x})$.

### 3.2.1 Sampling with Langevin dynamics

This work takes inspiration from the recent success of Langevin dynamics for sampling from unnormalized models (Du et al., 2019; Song et al., 2019a; Nijkamp et al., 2019b). In particular, we design an (overdamped) Langevin diffusion process whose stationary distribution is the noise distribution $p_{\text{m}}(\mathbf{x})$. Such a process can then be simulated by discretizing the time steps into intervals of length $\tau$. To draw a sample from $p_{\text{m}}(\mathbf{x})$, we first sample $\mathbf{x}_0$ from any prior distribution, then we simulate Langevin dynamics for $k = 0, \dots, K - 1$ steps according to the following discretized process:

$$\mathbf{x}_{k+1} = \mathbf{x}_k + \frac{\tau}{2} \nabla_{\mathbf{x}} \log p_{\text{m}}(\mathbf{x}_k) + \sqrt{\tau}\, \boldsymbol{\epsilon} \quad \text{where } \boldsymbol{\epsilon} \sim \mathcal{N}(\mathbf{0}, I). \tag{3.6}$$

In theory, to correct for the discretization error we typically need a Metropolis-Hastings step to ensure convergence [①]. However, in practice, if the number of steps $K \to \infty$ is sufficiently large and the discretization $\tau \to 0$ is small enough, this MH step can be ignored. The bulk of the complexity in Langevin dynamics arises from evaluation of the score term $\nabla_{\mathbf{x}} \log p_{\text{m}}(\mathbf{x})$. Fortunately, in the case of AdaNCE, the noise distribution $p_{\text{m}}(\mathbf{x})$ is a previous instance of the EBM $p_{\theta_t}(\mathbf{x}) = \frac{1}{Z_{\theta_t}} e^{-E_{\theta_t}(\mathbf{x})}$ at iteration $t$. As such, the score term for a Langevin diffusion step has a simple expression given by $\nabla_{\mathbf{x}} \log p_{\text{m}}(\mathbf{x}) = -\nabla_{\mathbf{x}} E_{\theta_t}(\mathbf{x})$.

Even with gradient-based MCMC methods like Langevin dynamics, running the sampler to convergence is computationally expensive. This is especially true for high-dimensional data where it may take upwards of several thousand steps for Langevin dynamics to achieve steady-state. As a result, we sometimes resort to using approximate

---

① Langevin dynamics with a MH step is also know as the Metropolis-Adjusted Langevin Algorithm (MALA).

sampling methods that are more efficient but not guaranteed to reach equilibrium; and this usually involves running the chain for fewer steps than required for convergence. One line of work that explores the effects of non-convergent MCMC is Nijkamp et al. (2019a). The authors find that, contrary to popular belief, the models learned using non-convergent maximum likelihood estimation are in fact easier to train and still provide high-quality synthesis. Following their experiments, we opt to train AdaNCE using non-convergent Langevin dynamics: we use step size $\tau = 1$, with a separate Brownian noise term with standard deviation $\sqrt{\tau} = 0.005$. The burn-in period for the sampler is $K = 100$ steps. One thing to note is that the models trained using non-convergent MCMC more resemble neural samplers and not approximations of the data density. This means that the learned energy landscape $E_\theta(\mathbf{x})$ acts as a truncated sampling path from which samples are generated, and that the EBM $p_\theta(\mathbf{x}) = \frac{1}{Z_\theta}e^{-E_\theta(\mathbf{x})}$ is not necessarily an approximation to $p_{\text{data}}(\mathbf{x})$. If we use convergent MCMC to sample $\mathbf{x}$ from a model trained using non-convergent methods, then $\mathbf{x}$ has a tendency to experience mode collapse.

Another important aspect of MCMC sampling is the initialization of the Markov chain. This choice of initial distribution not only impacts the convergence rate but also influences the exploration of different modes of the data. In particular, there are three popular techniques in which the Markov chain is initialized:

1. *Noise*: Initialize the Markov chain with samples from some noise distribution (typically uniform or Gaussian). The MCMC chains initialized in this way usually take longer to converge but are more likely to explore all modes of the data distribution.

2. *Data*: Initialize the Markov chain with samples from the data distribution. This method is introduced in contrastive divergence (CD) (Hinton, 2002). Data initialized Markov chains are faster to converge; however, models trained in this way have a tendency to learn the trivial solution $E_\theta(\mathbf{x}) = c$.

3. *Persistent*: Initialize the Markov chain from a sample buffer. This buffer stores previously generated samples and uses these past samples as starting points for the current chain. Persistent MCMC is asymptotically fast to converge but also has a tendency to learn the trivial solution $E_\theta(\mathbf{x}) = c$.

In order to alleviate the disadvantages of each method, Du et al. (2019) introduce persistent initialization with noise rejuvenation. The MCMC chains initialized in this way draw samples from a replay sample buffer 95% of the time and from some noise distribution otherwise. The idea is that by combining both noise and persistent initialization, we are able to ensure that the chain converges quickly while mitigating the trivial solution.

Following this line of thinking, the MCMC chains employed in AdaNCE use persistent initialization with noise rejuvenation. However, unlike in Du et al. (2019), we find that a noise rejuvenation rate of $\mathcal{R} = 5\%$ still leads to the trivial solution during training. Rather, through simple hyperparameter tuning, we discover that a rejuvenation rate of $\mathcal{R} = 10\% \sim 25\%$ works well. Furthermore, we find that the steady-state samples of the models trained using non-convergent MCMC with $\mathcal{R} = 10\% \sim 25\%$ typically experience less mode collapse.

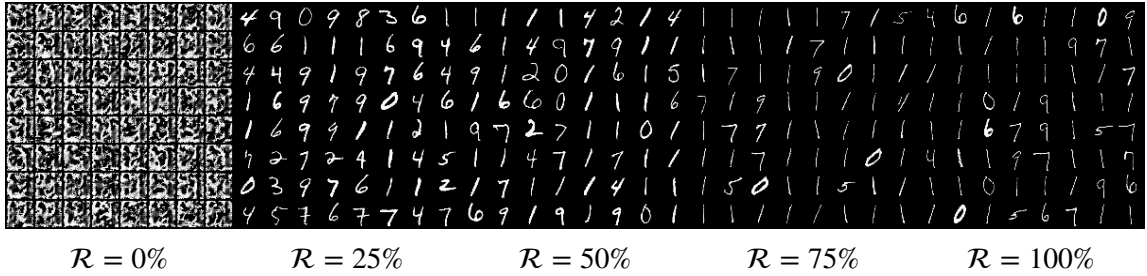

| $\mathcal{R} = 0\%$ | $\mathcal{R} = 25\%$ | $\mathcal{R} = 50\%$ | $\mathcal{R} = 75\%$ | $\mathcal{R} = 100\%$ |

Figure 3.1   Steady-state (long-run) samples of the models trained using persistent initialization with noise rejuvenation rate $\mathcal{R}$. At $\mathcal{R} = 0\%$ the model learns the trivial solution $E_\theta(\mathbf{x}) = c$. When $\mathcal{R} = 100\%$ (equivalent to noise-initialization), the steady-state samples experience mode collapse.

### 3.2.2   Spectral normalization for improved stability

Unfortunately, the AdaNCE algorithm in its current form is almost always guaranteed to experience training instability. The root cause of this instability lies in the fact that updates to the EBM have a tendency to cause the energy landscape $E_\theta(\mathbf{x})$ to have sharp changes in gradients. This makes sampling with Langevin dynamics numerically unstable since values of the score term $\nabla_{\mathbf{x}} \log p_\theta(\mathbf{x}) = -\nabla_{\mathbf{x}} E_\theta(\mathbf{x})$ may blow up. To further detail this phenomenon, we record the average gradient magnitude $v$ of the energy along a Langevin sample path $\mathbf{x}_0, \dots, \mathbf{x}_{K-1}$ during model training:

$$v = \mathbb{E}_{\mathbf{x}_0, \dots, \mathbf{x}_{K-1}} \left[ \frac{1}{K} \sum_{\ell=0}^{K-1} \left\| \nabla_{\mathbf{x}} E_\theta(\mathbf{x}_\ell) \right\|_2 \right].$$

The observations of the gradient magnitudes in Figure 3.2 tell us that regardless of the rejuvenation rate, $v$ seems to increase indefinitely during training. In fact, all these graphs record the gradient magnitudes just before $v$ jumps towards infinity, thereby causing the Langevin sampler to fail and training to collapse.

One way we can prevent steep gradients in the energy function is by constraining $E_\theta(\mathbf{x})$ to be $K$-Lipschitz continuous. Given metric spaces $(\mathcal{X}, d_{\mathcal{X}})$ and $(\mathcal{Y}, d_{\mathcal{Y}})$, the Lips-

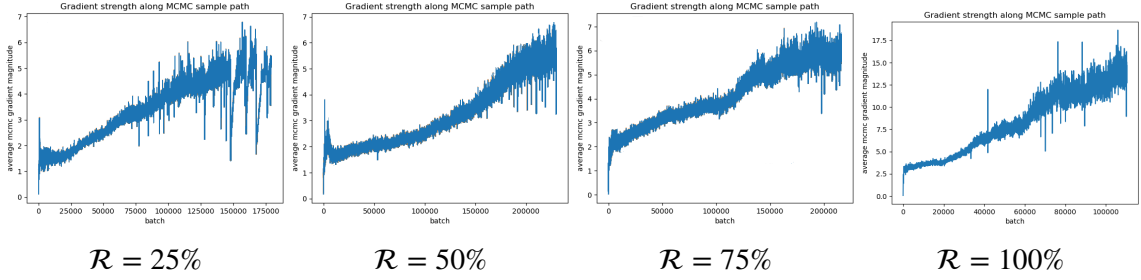

$$\mathcal{R} = 25\% \qquad \mathcal{R} = 50\% \qquad \mathcal{R} = 75\% \qquad \mathcal{R} = 100\%$$

Figure 3.2  Average magnitude $\nu$ of the score term along Langevin sample paths. These gradient magnitudes are recorded at every 10-th training iteration.

chitz norm $\|\cdot\|_{\text{Lip}}$ of a function $f : \mathcal{X} \to \mathcal{Y}$ is defined by

$$\|f\|_{\text{Lip}} := \sup_{x_1 \neq x_2} \frac{d_{\mathcal{Y}}(f(x_1), f(x_2))}{d_{\mathcal{X}}(x_1, x_2)}. \tag{3.7}$$

The function $f$ is then $K$-Lipschitz continuous if its Lipschitz norm is bounded above by $K$. In the case where $d_{\mathcal{X}}$ and $d_{\mathcal{Y}}$ are standard euclidean distances on $\mathbb{R}^m$ and $\mathbb{R}^n$, $f$ is $K$-Lipschitz continuous if

$$\|f\|_{\text{Lip}} := \sup_{x_1 \neq x_2} \frac{\|f(x_1) - f(x_2)\|_2}{\|x_1 - x_2\|_2} \leq K. \tag{3.8}$$

Intuitively, $K$-Lipschitz continuity implies that a continuous function has gradients which are bounded in magnitude by the Lipschitz constant $K$. It turns out that if $f : \mathbf{h}_{\text{in}} \to \mathbf{h}_{\text{out}}$ is a linear operator of the form $f(\mathbf{h}) = W\mathbf{h} + \mathbf{b}$, then the Lipschitz norm of $f$ has the following simple form: $\|f\|_{\text{Lip}} := \sup_{\mathbf{h}} \|\nabla f(\mathbf{h})\| = \sup_{\mathbf{h}} \|W\| = \|W\|$, where $\|W\|$ is the $L_2$ matrix norm of $W$ (a.k.a., *spectral norm*) given by:

$$\|W\| := \max_{\mathbf{h} \neq \mathbf{0}} \frac{\|W\mathbf{h}\|_2}{\|\mathbf{h}\|_2} = \max_{\|\mathbf{h}\|_2 \leq 1} \|W\mathbf{h}\|_2 = \sigma_{\max}(W). \tag{3.9}$$

The term $\sigma_{\max}(W)$ denotes the largest singular value of $W$.

Let us now consider a general feed-forward neural network $f(\mathbf{x}) = (f_L \circ \cdots \circ f_1)(\mathbf{x})$ with layers defined by $\mathbf{h}^{(\ell)} = f_\ell(\mathbf{h}^{(\ell-1)}) = \alpha^{(\ell)} (W^{(\ell)}\mathbf{h}^{(\ell-1)} + \mathbf{b}^{(\ell)})$. If we assume the Lipschitz norm of every activation $\|\alpha^{(\ell)}\|_{\text{Lip}}$ is equal to 1 (true for many activations), then we can use the property $\|f_1 \circ f_2\|_{\text{Lip}} \leq \|f_1\|_{\text{Lip}} \|f_2\|_{\text{Lip}}$ to determine a lower bound on the Lipschitz norm of the network:

$$\|f\|_{\text{Lip}} \leq \|f_L\|_{\text{Lip}} \cdots \|f_1\|_{\text{Lip}} \tag{3.10}$$
$$= \prod_{i=1}^{L} \|\alpha^{(i)}\|_{\text{Lip}} \|W^{(i)}\| = \prod_{i=1}^{L} \|W^{(i)}\|.$$

This bound tells us that by normalizing the spectral norm of the weights of each layer (i.e., $\overline{W} := \frac{W}{\|W\|}$), the neural network $f$ will be 1-Lipschitz continuous. This method of enforcing Lipschitz continuity is known as *spectral normalization* (Miyato et al., 2018).

In order to evaluate the spectral norm $\|W\| = \sigma_{\max}(W)$, we need to find the largest singular value of $W$. The naive approach to this problem involves applying singular value decomposition; however, this is computationally expensive, especially since the spectral norm needs to be computed for each layer in the network. Instead, we can use the *power iteration method* to approximate $\sigma_{\max}(W)$ with little extra training cost. The basic idea is to first sample a random vector $\bar{\mathbf{u}}$ from some isotropic distribution. Next, we can produce the left and right singular vectors according to $N$ iterations of the following update rule: $\bar{\mathbf{v}} \leftarrow W^\mathsf{T}\bar{\mathbf{u}}/\|W^\mathsf{T}\bar{\mathbf{u}}\|_2$ and $\bar{\mathbf{u}} \leftarrow W\bar{\mathbf{v}}/\|W\bar{\mathbf{v}}\|_2$. The resulting spectral norm can then be approximated by $\|W\| = \sigma_{\max}(W) \approx \bar{\mathbf{u}}^\mathsf{T} W \bar{\mathbf{v}}$. The basic spectral normalization procedure is presented in Algorithm 3.2. Miyato et al. (2018) further take advantage of the fact that the changes in $W$ at each update step are small, and therefore use the $\bar{\mathbf{u}}$ computed at each step as the initial vector in the subsequent step. It turns out that with this initialization procedure, one cycle ($N = 1$) of the power iteration method is sufficient for satisfactory spectral normalization.

---

**Algorithm 3.2** Spectral Normalization

**Input:** Network with weights $W_l \in \mathbb{R}^{m_l \times n_l}$ for layers $l = 1..L$

**Initialize:** $\bar{\mathbf{u}}_l \in \mathbb{R}^{m_l}$ with an isotropic distribution for each layer $l = 1..L$

1: **for** each layer $l = 1 : L$ **do**
2:    **for** $i = 1 : N$ **do**
3:       $\bar{\mathbf{v}}_l \leftarrow W_l^\mathsf{T}\bar{\mathbf{u}}_l/\|W_l^\mathsf{T}\bar{\mathbf{u}}_l\|_2$
4:       $\bar{\mathbf{u}}_l \leftarrow W_l\bar{\mathbf{v}}_l/\|W_l\bar{\mathbf{v}}_l\|_2$
5:    **end for**
6:    $\overline{W}_l \leftarrow W_l/\sigma_{\max}(W_l) \approx W_l/\bar{\mathbf{u}}_l^\mathsf{T} W \bar{\mathbf{v}}_l$
7:    $W_l \leftarrow W_l - \eta\nabla_{W_l}\mathcal{L}(\overline{W}_l)$
8: **end for**

---

Consequently, using the work of Miyato et al. (2018), we apply spectral normalization to all layers of the energy function $E_\theta(\mathbf{x})$. We find that this greatly stabilizes training and prevents the Langevin sampler from failing. Just like before, we record the average gradient magnitudes $v$ during training and find that spectral normalization regularizes these gradients along the sample path. The graphs of the sample gradient strength for $\mathcal{R} = 50\%$ and $\mathcal{R} = 100\%$ are shown in Figure 3.3, and the effects of spectral normalization become apparent when we compare these results to those of Figure 3.2. With spectral normalization, we can see that the gradient strength along MCMC sample paths stabilizes to an approximately constant value. In contrast, without any regularization, the gradient strength steadily increases until values become unstable.

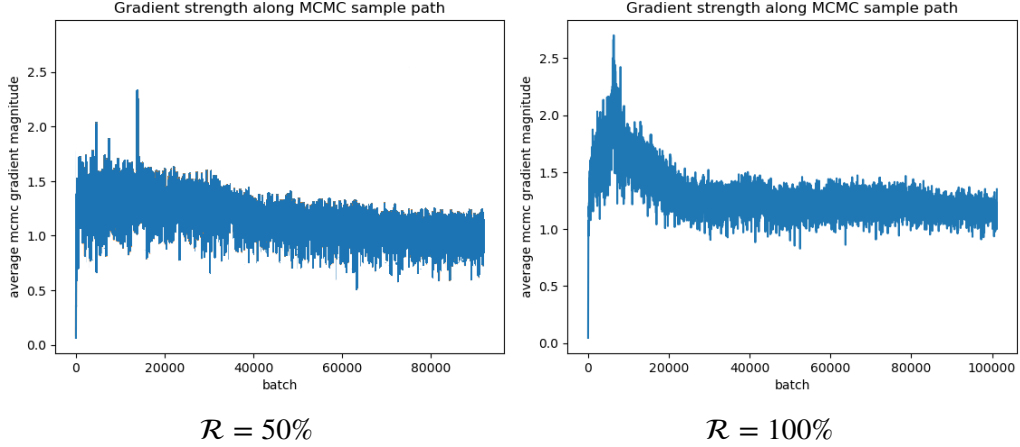

$$\mathcal{R} = 50\% \qquad\qquad\qquad \mathcal{R} = 100\%$$

Figure 3.3    Average magnitude $\nu$ of the score term along Langevin sample paths in a spectral normalized network. These gradient magnitudes are recorded at every 10-th training iteration.

Finally, we note that Nijkamp et al. (2019a) find that spectral normalization is not necessary for stable training. Their work instead relies on careful initialization and fine-tuning so that the gradient strength $\nu$ is approximately constant. This makes the training process much more complex, and we argue that spectral normalization is a better alternative. On the other hand, Du et al. (2019) are some of the first authors to apply spectral normalization to EBMs; however, unlike us they did not elaborate on this phenomenon in detail.

## 3.3   Connection to Maximum Likelihood Estimation

Now that we have explained the details of our algorithm, let us instead analyze some of its theoretical properties. We find that, remarkably, there is an unlikely connection between self-adapting noise-contrastive estimation and maximum likelihood estimation for EBM training. In particular, when the noise distribution is tuned at every iteration (i.e., $\mathcal{K} = 1$), the AdaNCE objective corresponds to MLE. To see why this is the case, we can consider the gradient of the AdaNCE objective (3.5) given by:

$$\nabla_\theta J(\theta) = \nabla_\theta \left( -\mathbb{E}_{p_{\mathrm{m}}(\mathbf{x})} \left[ \log \frac{p_{\mathrm{m}}(\mathbf{x})}{p_{\mathrm{m}}(\mathbf{x}) + p_\theta(\mathbf{x})} \right] - \mathbb{E}_{p_{\mathrm{data}}(\mathbf{x})} \left[ \log \frac{p_\theta(\mathbf{x})}{p_{\mathrm{m}}(\mathbf{x}) + p_\theta(\mathbf{x})} \right] \right)$$

$$= \mathbb{E}_{p_{\mathrm{m}}(\mathbf{x})} \left[ \frac{1}{p_{\mathrm{m}}(\mathbf{x}) + p_\theta(\mathbf{x})} \nabla_\theta p_\theta(\mathbf{x}) \right] - \mathbb{E}_{p_{\mathrm{data}}(\mathbf{x})} \left[ \frac{p_{\mathrm{m}}(\mathbf{x})}{p_\theta(\mathbf{x})} \frac{1}{p_{\mathrm{m}}(\mathbf{x}) + p_\theta(\mathbf{x})} \nabla_\theta p_\theta(\mathbf{x}) \right]$$

$$= \mathbb{E}_{p_m(\mathbf{x})} \left[ \frac{p_\theta(\mathbf{x})}{p_m(\mathbf{x}) + p_\theta(\mathbf{x})} \nabla_\theta \log p_\theta(\mathbf{x}) \right] - \mathbb{E}_{p_{\text{data}}(\mathbf{x})} \left[ \frac{p_m(\mathbf{x})}{p_m(\mathbf{x}) + p_\theta(\mathbf{x})} \nabla_\theta \log p_\theta(\mathbf{x}) \right].$$

If the adaptive interval is $\mathcal{K} = 1$, then the noise distribution $p_m(\mathbf{x})$ is set to be the updated model distribution $p_\theta(\mathbf{x})$ at every iteration. As such, at every step of the algorithm we have $p_m(\mathbf{x}) \equiv p_\theta(\mathbf{x})$ and we can simplify $\nabla_\theta J(\theta)$ to be:

$$\nabla_\theta J(\theta) = \frac{1}{2} \mathbb{E}_{p_\theta(\mathbf{x})} \left[ \nabla_\theta \log p_\theta(\mathbf{x}) \right] - \frac{1}{2} \mathbb{E}_{p_{\text{data}}(\mathbf{x})} \left[ \nabla_\theta \log p_\theta(\mathbf{x}) \right].$$

In the case where the model is an EBM governed by the Boltzmann distribution $p_\theta(\mathbf{x}) = \frac{1}{Z_\theta} e^{-E_\theta(\mathbf{x})}$, we can use the fact that $\nabla_\theta \log p_\theta(\mathbf{x}) = -\nabla_\theta E_\theta(\mathbf{x}) - \nabla_\theta \log Z_\theta$ to further simplify this gradient objective into:

$$\nabla_\theta J(\theta) = \frac{1}{2} \underbrace{\left( \mathbb{E}_{p_{\text{data}}(\mathbf{x})} \left[ \nabla_\theta E_\theta(\mathbf{x}) \right] - \mathbb{E}_{p_\theta(\mathbf{x})} \left[ \nabla_\theta E_\theta(\mathbf{x}) \right] \right)}_{-\nabla_\theta J_{\text{MLE}}(\theta)},$$

where the expression in the parenthesis is equivalent to the maximum likelihood objective (2.4). Therefore, we can see that when $\mathcal{K} = 1$, minimizing the AdaNCE objective is identical (up to a scalar factor) to maximizing the likelihood of EBMs.

## 3.4 Discussion

We conclude with a brief summary of the key points of this chapter. While noise-contrastive estimation is a powerful method for learning probabilistic models, it suffers from the curse of dimensionality. One line of work that aims to improve the scalability of NCE is by utilizing an adaptive noise distribution which dynamically adjusts to closer approximate the target distribution during training. However, current methods exploring this idea rely on modelling the noise distribution with a separate generative model. Hence, this noise distribution needs to be concurrently trained with our model distribution.

As a way of mitigating this extra computational complexity, we instead propose to use static instances of the model distribution along its training trajectory as the noise distribution. This noise distribution automatically adapts to be closer to the target distribution, and is aptly referred to as a self-adapting noise distribution. Noise-contrastive estimation that features this self-adapting noise is then referred to as self-adapting noise-contrastive estimation. Consequently, training a probabilistic model using AdaNCE involves several NCE steps followed by a single noise update step.

There are several key details of AdaNCE worth mentioning. Firstly, the NCE steps require us to sample from the noise distribution. Since this noise distribution is complex and dynamic, we choose to rely on short-run Langevin dynamics to approximately sample from this noise. Secondly, naive execution of AdaNCE results in sharp changes in the model's score function. This causes training instability when using Langevin dynamics to draw samples. In order to mitigate this issue, we applied spectral normalization to all layers of our model to bind the model's Lipschitz norm.

Finally, we note that AdaNCE is identical to MLE when the adaptive interval is $\mathcal{K} = 1$. This is because the gradients of both objectives are equal. In fact, the following chapter explores generalizing AdaNCE under the framework of Bregman divergences. It turns out that this generalization also reduces to MLE in the case when $\mathcal{K} = 1$.

# CHAPTER 4   GENERALIZATION UNDER THE BREGMAN DIVERGENCE

This chapter explores generalizing AdaNCE under the framework of Bregman divergences. First, we explain what Bregman divergences are and how NCE is an instance of Bregman divergence minimization. Next, we show that this framework can also be used to explain AdaNCE, and we refer to this generalization as self-adapting Bregman ratio matching (AdaBRM). Finally, we show that, just like AdaNCE, AdaBRM also reduces to MLE in certain cases.

## 4.1   What are Bregman Divergences?

Bregman divergences are a measure of discrepancy between two points, usually defined with a strictly convex function $\Psi$. They can be thought of as a generalization of the Euclidean distance to a class of divergences with similar properties. In particular, the Bregman divergence $d_\Psi(\mathbf{x}, \mathbf{y})$ between two points $\mathbf{x}, \mathbf{y}$ associated with a continuous, strictly convex function $\Psi : \Omega \to \mathbb{R}$ is the difference between the value $\Psi(\mathbf{x})$ and the value of the first-order Taylor expansion of $\Psi(\mathbf{x})$ around point $\mathbf{y}$:

$$d_\Psi(\mathbf{x}, \mathbf{y}) = \Psi(\mathbf{x}) - \Psi(\mathbf{y}) - \langle \nabla\Psi(\mathbf{y}), \mathbf{x} - \mathbf{y} \rangle. \tag{4.1}$$

We note that because $\Psi$ is strictly convex and differentiable, the Bregman divergence is non-negative for all points $\mathbf{x}, \mathbf{y}$ in the domain of $\Psi$. Furthermore, this divergence is minimized at $d_\Psi(\mathbf{x}, \mathbf{y}) = 0$ only when $\mathbf{x} = \mathbf{y}$.

Although the Bregman divergence is typically defined between two points, we can extend the concept to encompass functions as well. If $\mathbf{f}$ and $\mathbf{g}$ are two vector-valued functions, we can define the Bregman divergence $D_\Psi(\mathbf{f}, \mathbf{g})$ between these functions as the sum of all Bregman divergences between points in the domain, possibly weighted by some measure $\mu$:

$$D_\Psi(\mathbf{f}, \mathbf{g}) = \int d_\Psi(\mathbf{f}, \mathbf{g}) \, \mathrm{d}\mu$$
$$= \int \Psi(\mathbf{f}) - \Psi(\mathbf{g}) - \nabla\Psi(\mathbf{g})^\top (\mathbf{f} - \mathbf{g}) \, \mathrm{d}\mu. \tag{4.2}$$

Since $D_\Psi(\mathbf{f}, \mathbf{g})$ is a valid divergence, we can approximate a fixed target function $\mathbf{f}$ with a variational function $\mathbf{g}$ by minimizing the Bregman divergence with respect to $\mathbf{g}$. This learning procedure is equivalent to minimizing the following Bregman loss:

$$L_{\Psi}(\mathbf{g}) = \int -\Psi(\mathbf{g}) - \nabla\Psi(\mathbf{g})^{\top}(\mathbf{f} - \mathbf{g}) \, d\mu. \tag{4.3}$$

Following the work of Gutmann and Hirayama (2012), we can define an alternative parameterization of the Bregman loss in the case where $f$ and $g$ are scalar-valued functions. Since most of the functions we work with are scalar densities, this parameterization will prove very useful later on. We can describe this parameterization as follows:

$$S_0(g) = -\Psi(g) + \Psi'(g)g \tag{4.4}$$

$$S_1(g) = \Psi'(g), \tag{4.5}$$

where by definition and the fact that $\Psi$ is strictly convex, these functions must satisfy

$$\frac{S_0'(g)}{S_1'(g)} = g \qquad \text{and} \qquad S_1'(g) > 0. \tag{4.6}$$

Using this parameterization, we can now express the Bregman loss (4.3) between two scalar-valued functions $f, g$ as a much simpler expression:

$$L_S(g) = \int S_0(g) - S_1(g)f \, d\mu. \tag{4.7}$$

We conclude this section by noting that for specific choices of the convex function $\Psi$, the Bregman loss reduces to several commonly seen statistical distances. For instance, if $\Sigma$ is a positive definite covariance matrix and $\Psi(\mathbf{u}) = \frac{1}{2}\mathbf{u}^{\top}\Sigma\mathbf{u}$, then the Bregman loss is equivalent to the squared Mahalanobis distance $L_{\Psi}(\mathbf{g}) = \int \frac{1}{2}(\mathbf{f} - \mathbf{g})^{\top}\Sigma(\mathbf{f} - \mathbf{g}) \, d\mu + \text{const}$. Furthermore, when $\Psi(u) = u \log u$ we get the generalized Kullback–Leibler divergence $L_{\Psi}(g) = \int f \log \frac{f}{g} \, d\mu - \int f \, d\mu + \int g \, d\mu + \text{const}$. As a matter of fact, in the following section we show that noise-contrastive estimation can also be described in the framework of Bregman divergences.

## 4.2 Bregman Ratio Matching

It turns out that noise-contrastive estimation is a special case of Bregman divergence minimization. In particular, it is an instance of *density-ratio matching under the framework of Bregman divergences* (Sugiyama et al., 2011; Gutmann et al., 2012). To see why this is the case, let us consider the alternate Bregman loss (4.7) where $f(\mathbf{x}) = \frac{p_{\text{data}}(\mathbf{x})}{p_{\text{n}}(\mathbf{x})}$ is the target density ratio of data-to-noise, and $g(\mathbf{x}) = \frac{\tilde{p}_{\theta}(\mathbf{x})}{p_{\text{n}}(\mathbf{x})}$ is the variational density ratio of model-to-noise. Here we explicitly indicate that $\tilde{p}_{\theta}(\mathbf{x})$ is an unnormalized model. If we let $\mu$ be the probability measure associated with the cdf of $p_{\text{n}}(\mathbf{x})$, then the Bregman loss takes on the following simple form:

$$L_{\text{BRM}}(\theta) = \int \left( S_0(g(\mathbf{x})) - S_1(g(\mathbf{x}))f(\mathbf{x}) \right) p_{\text{n}}(\mathbf{x})\, d\mathbf{x} \qquad (4.8)$$

$$= \mathbb{E}_{p_{\text{n}}(\mathbf{x})} \left[ S_0(g(\mathbf{x})) \right] - \mathbb{E}_{p_{\text{data}}(\mathbf{x})} \left[ S_1(g(\mathbf{x})) \right].$$

We denote this expression as the *Bregman ratio matching* (BRM) objective. Unbiased estimates of this objective are easily obtained as long as we can sample from $p_{\text{data}}$ and $p_{\text{n}}$. If the corresponding convex function is $\Psi(u) = u \log u - (1+u)\log(1+u)$, or equivalently when $S_0(u) = \log(1+u)$ and $S_1(u) = \log(u) - \log(1+u)$, then the Bregman ratio matching objective can be expressed as:

$$L_{\text{BRM}}(\theta) = \mathbb{E}_{p_{\text{n}}(\mathbf{x})} \left[ \log(1 + g(\mathbf{x})) \right] - \mathbb{E}_{p_{\text{data}}(\mathbf{x})} \left[ \log \frac{g(\mathbf{x})}{1 + g(\mathbf{x})} \right]$$

$$= \underbrace{- \mathbb{E}_{p_{\text{n}}(\mathbf{x})} \left[ \log \frac{p_{\text{n}}(\mathbf{x})}{p_{\text{n}}(\mathbf{x}) + \tilde{p}_\theta(\mathbf{x})} \right] - \mathbb{E}_{p_{\text{data}}(\mathbf{x})} \left[ \log \frac{\tilde{p}_\theta(\mathbf{x})}{p_{\text{n}}(\mathbf{x}) + \tilde{p}_\theta(\mathbf{x})} \right]}_{J_{\text{NCE}}(\theta)}.$$

This form of BRM is identical to the noise-contrastive estimation objective (2.13); hence, this framework also provides an explanation for why NCE learns the model's normalization constant: by matching the density ratios $f = \frac{p_{\text{data}}}{p_{\text{n}}}$ with $g = \frac{\tilde{p}_\theta}{p_{\text{n}}}$, we indirectly fit $\tilde{p}_\theta(\mathbf{x})$ to $p_{\text{data}}(\mathbf{x})$. Therefore, the optimal model parameters $\theta^*$ will always learn a partition function such that $\tilde{p}_{\theta^*}(\mathbf{x}) = e^{-E_{\theta^*}(\mathbf{x})}$ integrates to one.

## 4.3   Self-Adapting Bregman Ratio Matching

In this section, we show that AdaNCE is also an instance of density-ratio matching with Bregman divergences. Specifically, when both density ratios $f$ and $g$ feature a self-adapting noise distribution $p_{\text{m}}(\mathbf{x})$ [①], we arrive at an objective that generalizes AdaNCE. First, let us consider replacing $p_{\text{n}}$ with $\tilde{p}_{\text{m}}$ in the density ratios $f$ and $g$ of the previous section. We use the unnormalized $p_{\text{m}}$ since, by definition, the self-adapting noise distribution is only known up to a normalization constant. Therefore, we now consider the alternate Bregman loss (4.7) associated with matching $f = \frac{p_{\text{data}}}{\tilde{p}_{\text{m}}}$ and $g = \frac{\tilde{p}_\theta}{\tilde{p}_{\text{m}}}$. If we let $\mu(d\mathbf{x}) = p_{\text{m}}(\mathbf{x})\, d\mathbf{x}$, then this Bregman loss is given by:

$$L_{\text{S}}(\theta) = \int \left( S_0(g(\mathbf{x})) - S_1(g(\mathbf{x})) \frac{p_{\text{data}}(\mathbf{x})}{\tilde{p}_{\text{m}}(\mathbf{x})} \right) p_{\text{m}}(\mathbf{x})\, d\mathbf{x} \qquad (4.9)$$

---

[①] This self-adapting noise distribution is a static instance of the trained EBM at a previous iteration. It is updated the same way as in Algorithm 3.1.

$$= \int S_0(g(\mathbf{x})) p_{\mathrm{m}}(\mathbf{x}) \, d\mathbf{x} - \int \frac{1}{Z_{\mathrm{m}}} S_1(g(\mathbf{x})) p_{\mathrm{data}}(\mathbf{x}) \, d\mathbf{x}$$

$$= \mathbb{E}_{p_{\mathrm{m}}(\mathbf{x})} \left[ S_0(g(\mathbf{x})) \right] - \mathbb{E}_{p_{\mathrm{data}}(\mathbf{x})} \left[ \frac{1}{Z_{\mathrm{m}}} S_1(g(\mathbf{x})) \right].$$

Unfortunately, estimation of this expression is extremely challenging due to the intractable partition function $Z_{\mathrm{m}}$. Instead, we can consider matching the following density ratios $f = \frac{p_{\mathrm{data}}}{p_{\mathrm{m}}}$ and $g = \frac{\tilde{p}_\theta}{\tilde{p}_{\mathrm{m}}}$, where the target density ratio deals with the normalized $p_{\mathrm{m}}(\mathbf{x})$ and the variational density ratio deals with the unnormalized $\tilde{p}_{\mathrm{m}}(\mathbf{x})$. Once again we let the weighted measure be $\mu(d\mathbf{x}) = p_{\mathrm{m}}(\mathbf{x}) \, d\mathbf{x}$. In this case, the Bregman divergence objective takes on a much simpler form:

$$L_{\mathrm{AdaBRM}}(\theta) = \int \left( S_0(g(\mathbf{x})) - S_1(g(\mathbf{x})) \frac{p_{\mathrm{data}}(\mathbf{x})}{p_{\mathrm{m}}(\mathbf{x})} \right) p_{\mathrm{m}}(\mathbf{x}) \, d\mathbf{x} \tag{4.10}$$

$$= \mathbb{E}_{p_{\mathrm{m}}(\mathbf{x})} \left[ S_0(g(\mathbf{x})) \right] - \mathbb{E}_{p_{\mathrm{data}}(\mathbf{x})} \left[ S_1(g(\mathbf{x})) \right].$$

We denote this expression as the *self-adapting Bregman ratio matching* (AdaBRM) objective. Unlike (4.9), this AdaBRM objective is straightforward to estimate using Monte Carlo sampling. In particular, when $S_0(u) = \log(1+u)$ and $S_1(u) = \log(u) - \log(1+u)$, or equivalently when $\Psi(u) = u \log(u) - (1+u) \log(1+u)$, this self-adapting BRM objective corresponds to AdaNCE:

$$L_{\mathrm{AdaBRM}}(\theta) = \mathbb{E}_{p_{\mathrm{m}}(\mathbf{x})} \left[ \log(1 + g(\mathbf{x})) \right] - \mathbb{E}_{p_{\mathrm{data}}(\mathbf{x})} \left[ \log \frac{g(\mathbf{x})}{1 + g(\mathbf{x})} \right] \tag{4.11}$$

$$= \underbrace{- \mathbb{E}_{p_{\mathrm{m}}(\mathbf{x})} \left[ \log \frac{\tilde{p}_{\mathrm{m}}(\mathbf{x})}{\tilde{p}_{\mathrm{m}}(\mathbf{x}) + \tilde{p}_\theta(\mathbf{x})} \right] - \mathbb{E}_{p_{\mathrm{data}}(\mathbf{x})} \left[ \log \frac{\tilde{p}_\theta(\mathbf{x})}{\tilde{p}_{\mathrm{m}}(\mathbf{x}) + \tilde{p}_\theta(\mathbf{x})} \right]}_{J_{\mathrm{AdaNCE}}(\theta)}.$$

Consequently, we have shown that self-adapting noise-contrastive estimation is also an instance of density-ratio matching with Bregman divergences. In particular, it corresponds to matching $f = \frac{p_{\mathrm{data}}}{p_{\mathrm{m}}}$ with $g = \frac{\tilde{p}_\theta}{\tilde{p}_{\mathrm{m}}}$. Moreover, AdaBRM provides us with an explanation for why we can use unnormalized values of $p_{\mathrm{m}}(\mathbf{x})$ in the AdaNCE objective (3.5): this is because matching the density ratios $f$ and $g$ will implicitly match $\tilde{p}_\theta(\mathbf{x})$ to a scaled $p_{\mathrm{data}}(\mathbf{x})$, which consequently matches $p_\theta(\mathbf{x})$ to $p_{\mathrm{data}}(\mathbf{x})$. Hence, the optimal model will still learn the data distribution, with the only caveat being that the model's partition function is not learned. Fortunately, even though $Z_\theta$ is not learned, we only require the score function $\nabla_{\mathbf{x}} \log p_\theta(\mathbf{x})$ (which does not depend on the partition function) in order to generate new samples via gradient based MCMC.

## 4.4 Connection to Maximum Likelihood Estimation

To conclude our analysis, we show that, in the context of training EBMs, any flavor (any choice of convex $\Psi$) of self-adapting Bregman ratio matching reduces to maximum likelihood estimation when the adaptive noise distribution $p_{\mathrm{m}}(\mathbf{x})$ is updated at every training iteration (i.e., $\mathcal{K} = 1$). To see why this is the case, let us consider the AdaBRM objective (4.10) with $f = \frac{p_{\mathrm{data}}}{p_{\mathrm{m}}}$ and $g_\theta = \frac{\tilde{p}_\theta}{\tilde{p}_{\mathrm{m}}}$. If we take gradients of this objective we get:

$$\nabla_\theta L_{\mathrm{AdaBRM}}(\theta) = \nabla_\theta \left( \mathbb{E}_{p_{\mathrm{m}}(\mathbf{x})} \left[ S_0(g_\theta(\mathbf{x})) \right] - \mathbb{E}_{p_{\mathrm{data}}(\mathbf{x})} \left[ S_1(g_\theta(\mathbf{x})) \right] \right) \tag{4.12}$$

$$= \mathbb{E}_{p_{\mathrm{m}}(\mathbf{x})} \left[ S_0'(g_\theta(\mathbf{x})) \nabla_\theta g_\theta(\mathbf{x}) \right] - \mathbb{E}_{p_{\mathrm{data}}(\mathbf{x})} \left[ S_1'(g_\theta(\mathbf{x})) \nabla_\theta g_\theta(\mathbf{x}) \right]$$

$$= \mathbb{E}_{p_{\mathrm{m}}(\mathbf{x})} \left[ S_0'(g_\theta(\mathbf{x})) \frac{\nabla_\theta \tilde{p}_\theta(\mathbf{x})}{\tilde{p}_{\mathrm{m}}(\mathbf{x})} \right] - \mathbb{E}_{p_{\mathrm{data}}(\mathbf{x})} \left[ S_1'(g_\theta(\mathbf{x})) \frac{\nabla_\theta \tilde{p}_\theta(\mathbf{x})}{\tilde{p}_{\mathrm{m}}(\mathbf{x})} \right]$$

$$= \mathbb{E}_{p_{\mathrm{m}}(\mathbf{x})} \left[ S_0'(g_\theta(\mathbf{x})) g_\theta(\mathbf{x}) \nabla_\theta \log \tilde{p}_\theta(\mathbf{x}) \right]$$

$$\quad - \mathbb{E}_{p_{\mathrm{data}}(\mathbf{x})} \left[ S_1'(g_\theta(\mathbf{x})) g_\theta(\mathbf{x}) \nabla_\theta \log \tilde{p}_\theta(\mathbf{x}) \right].$$

When $\mathcal{K} = 1$, we update the noise distribution $p_{\mathrm{m}}(\mathbf{x})$ to be the current model distribution $p_\theta(\mathbf{x})$ at every training iteration. As a result, $p_{\mathrm{m}}(\mathbf{x}) = p_\theta(\mathbf{x})$ and the variational density ratio is $g_\theta(\mathbf{x}) = 1$. If we also use the fact that $S_0'(u) = u S_1'(u)$ (property 4.6), we can simplify the gradient objective to be:

$$\nabla_\theta L_{\mathrm{AdaBRM}}(\theta) = \mathbb{E}_{p_\theta(\mathbf{x})} \left[ S_0'(1) \nabla_\theta \log \tilde{p}_\theta(\mathbf{x}) \right] - \mathbb{E}_{p_{\mathrm{data}}(\mathbf{x})} \left[ S_1'(1) \nabla_\theta \log \tilde{p}_\theta(\mathbf{x}) \right]$$

$$= S_0'(1) \left( \mathbb{E}_{p_\theta(\mathbf{x})} \left[ \nabla_\theta \log \tilde{p}_\theta(\mathbf{x}) \right] - \mathbb{E}_{p_{\mathrm{data}}(\mathbf{x})} \left[ \nabla_\theta \log \tilde{p}_\theta(\mathbf{x}) \right] \right)$$

$$= S_0'(1) \underbrace{\left( -\mathbb{E}_{p_\theta(\mathbf{x})} \left[ \nabla_\theta E_\theta(\mathbf{x}) \right] + \mathbb{E}_{p_{\mathrm{data}}(\mathbf{x})} \left[ \nabla_\theta E_\theta(\mathbf{x}) \right] \right)}_{-\nabla_\theta J_{\mathrm{MLE}}(\theta)}.$$

Therefore, in regards to EBM learning, we can see that the training gradients of AdaBRM are equivalent to the scaled gradients of the maximum likelihood objective (2.4). This scaling factor is determined by the choice of the convex function $\Psi$.

## 4.5 Discussion

We conclude this chapter by summarizing the relationships between several key learning objectives. Firstly, we note that Bregman ratio matching is a generalization of noise-contrastive estimation. Specifically, when we use Bregman divergences to match the data-noise $\frac{p_{\mathrm{data}}}{p_{\mathrm{n}}}$ and model-noise $\frac{\tilde{p}_\theta}{p_{\mathrm{n}}}$ density ratios, we recover the NCE objective when the convex function is $\Psi(u) = u \log(u) - (1 + u) \log(1 + u)$. If we instead consider a self-adapting noise distribution $p_\mathrm{m}$, then matching $\frac{p_\mathrm{data}}{p_\mathrm{m}}$ to $\frac{\tilde{p}_\theta}{\tilde{p}_\mathrm{m}}$ using Bregman divergences corresponds to AdaBRM. It turns out that AdaBRM also generalizes AdaNCE, and the two objectives are equivalent when the convex function is again $\Psi(u) = u\log(u) - (1+u)\log(1+u)$. Finally, we show that regardless of the choice of convex function $\Psi$, the gradients of the AdaBRM objective are equivalent (up to a scalar factor) to the gradients of the MLE objective.

# CHAPTER 5 EXPERIMENTS & RESULTS

This chapter details the evaluation of self-adapting NCE for training energy-based models. We primarily judge the effectiveness of this algorithm on the synthesis quality of the EBM trained on a variety of datasets. First, we describe the experiments and key design decisions. Next, we present both the qualitative and quantitative results of these experiments. We conclude this chapter by summarizing the results and elaborating on key findings.

## 5.1 Experiments

The performance of AdaNCE is evaluated on image generation tasks. Specifically, we train an EBM on a multitude of image datasets including MNIST (Deng, 2012), CIFAR-10 (Krizhevsky, 2012), and CelebA (Liu et al., 2015). Furthermore, these experiments use a range of adaptive intervals $\mathcal{K}$ in order to examine the impact of this hyperparameter on training.

### 5.1.1 Model architecture

Since the data we want to synthesize is image-based, the natural design choice for the energy potential $E_\theta(\mathbf{x})$ follows a convolutional architecture. Previous authors have found that the vanilla ConvNet works well for training 32x32 images; as such, we tune their designs for our experiments. The following Table 5.1 details the architectures of the energy potentials used in our experiments. We specify convolution layers according to (kernel size, stride, padding) × output channels, and fully-connected layers are specified by their output dimension. The leaky ReLU activation was used for all architectures. Furthermore, we apply spectral normalization to all layers of the model.

### 5.1.2 Training details & hyperparameters

For the non-convergent Langevin sampler, we use a burn-in period of 100 steps, with a gradient term step-size of 1 and noise term standard deviation of 0.005. We initialize the MCMC chain with a persistent sample buffer with a noise rejuvenation rate of 25%. The buffer size is set to be 10000 samples. We scale images to the range (-1, 1), and train the model using the Adam optimizer with $\beta_1 = 0.0$ and $\beta_2 = 0.999$ with a learning rate of $10^{-4}$. We use a batch size of 128. Model specifications are given in Table 5.1, and we apply 1 cycle of the power iteration method for spectral normalization on all models. MNIST models are trained on 1 GPU for 1 day, and CIFAR-10 and CelebA models are trained over 2 days. We train several models using adaptive intervals in the range $\{1, 5, 10, 50\}$. These training details are summarized below:

- Langevin burn-in 100
- Langevin step size 1
- Langevin noise standard deviation 0.005
- Persistent sample buffer size 10000
- Noise rejuvenation rate 25%
- Normalize image to (-1,1)
- Train using Adam optimizer with $\beta_1 = 0.0$ and $\beta_2 = 0.999$
- Learning rate $10^{-4}$
- Batch size 128
- Spectral normalization with number of power iteration cycles 1
- Adaptive intervals $\{1, 5, 10, 50\}$

| Model | A | B | C |
|---|---|---|---|
| Input Size | (1, 28, 28) | (3, 32, 32) | |
| Convolution Layers | (3, 1, 1) x 64 | | |
| | (4, 2, 1) x 64 | | (4, 2, 1) x 128 |
| | (4, 2, 1) x 128 | | (4, 2, 1) x 256 |
| | (3, 2, 1) x 256 | (4, 2, 1) x 256 | (4, 2, 1) x 512 |
| | (4, 1, 0) x 512 | | (4, 1, 0) x 1024 |
| Linear Layers | Flatten | | |
| | FC-256 | | FC-512 |
| | FC-1 | | FC-256 |
| | | | FC-1 |

Table 5.1    Architectures for ConvNet energy potential $E_\theta(\mathbf{x})$ based on input sizes corresponding to different datasets. Model A is for MNIST; models B and C are for CIFAR and CelebA.

### 5.1.3 Evaluation metrics

We evaluate the performance of our self-adapting noise-contrastive estimation algorithm both qualitatively and quantitatively. While qualitative evaluation typically involves visual judgement, quantitative evaluation of generative models relies on several metrics; some more popular metrics include log-likelihood, Inception score (IS) (Salimans et al., 2016a), and Frechet inception distance (FID) (Heusel et al., 2017).

The de-facto way we can evaluate generative models is with the *log-likelihood*. In fact, it is often by maximizing this log-likelihood function (or minimizing the KL divergence) that models are trained. As such, this function provides a logical evaluation metric for how well a model is trained. Given a data distribution $p_{\text{data}}(\mathbf{x})$ and model distribution $p_\theta(\mathbf{x})$, the expected log-likelihood metric is given by:

$$\text{Log-Likelihood} := \mathbb{E}_{p_{\text{data}}(\mathbf{x})}\left[\log p_\theta(\mathbf{x})\right]. \tag{5.1}$$

Higher log-likelihood scores imply the model is more likely to synthesize samples of the data distribution. For explicit density models where the likelihood function is known, log-likelihoods are easy to evaluate. However, for implicit (and partially implicit) models like GANs and EBMs, the explicit density is often unknown or intractable. Thus the log-likelihood of these models requires estimation. In the context of EBMs, in order to evaluate the log-likelihood, we must estimate the intractable partition function $Z_\theta = \int e^{-E_\theta(\mathbf{x})}$. One way we can do this is by using annealed importance sampling (AIS) (Neal, 2001) to obtain a lower bound on the partition function. Unfortunately, in practice, these estimates are still largely unreliable for high-dimensional data; Du et al. (2019) found that when using AIS to estimate partition functions for CIFAR-10, the chain takes far too long to get any meaningful estimates. In particular, they ran AIS for over two days and still had very large discrepancies between the lowest and highest partition function estimates. Consequently, we choose to avoid using the log-likelihood as a metric in this thesis.

The *Inception score* is an evaluation metric based on our intuitive understanding of how an optimal generative model should perform. Unlike the log-likelihood, the IS can be determined for implicit models. The basic idea behind the Inception score is as follows: Suppose we have a classifier $f(\mathbf{x}) := p(\cdot|\mathbf{x})$ trained on some labelled image dataset like ImageNet. Let $p(y|\mathbf{x})$ denote the conditional class probability associated with image $\mathbf{x}$ where $y$ is the class label. Intuitively, generated images should have a conditional class distribution $p(y|\mathbf{x})$ with low entropy; this implies that images are distinct.

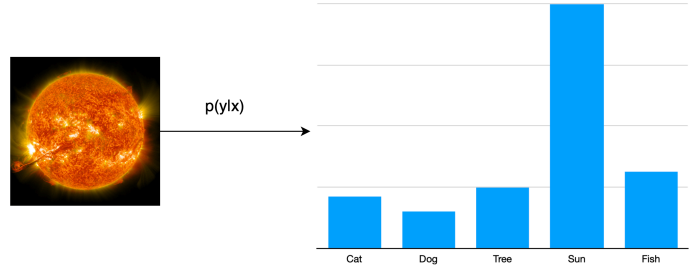

Figure 5.1    conditional class distribution $p(y|\mathbf{x})$ for an image of a sun $\mathbf{x}$.

On the other hand, synthesized data should also have a variety of different classes. This suggests that the marginal distribution $p_\theta(y) = \int p(y|\mathbf{x})p_\theta(\mathbf{x})\,\mathrm{d}\mathbf{x}$ should have high entropy.

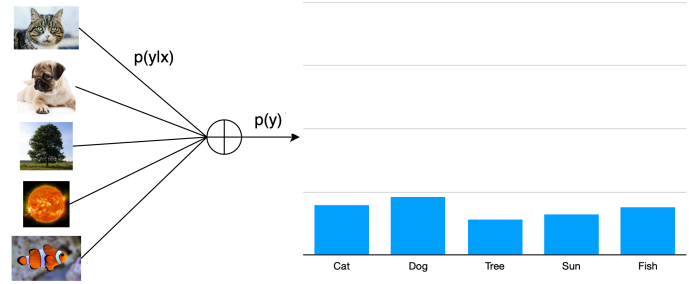

Figure 5.2    marginal class distribution $p_\theta(y)$ for the synthesized dataset $X$.

Therefore, one way we can evaluate if generated samples are both distinct and diverse is by comparing the conditional class distribution $p(y|\mathbf{x})$ with the marginal class distribution $p_\theta(y)$. In the inception score, this is done by determining the KL divergence between these distributions, with higher divergence signifying better synthesis quality. The log-IS is typically evaluated as the following:

$$
\begin{aligned}
\log \mathrm{IS} &= \mathbb{E}_{p_\theta(\mathbf{x})}\left[\mathrm{KL}(p(y|\mathbf{x})\|p(y))\right] \qquad\qquad (5.2)\\
&= \mathbb{E}_{p_\theta(\mathbf{x},y)}\left[\log p(y|\mathbf{x}) - \log p_\theta(y)\right]\\
&= \mathbb{E}_{p_\theta(\mathbf{x},y)}\left[\log p(y|\mathbf{x})\right] - \mathbb{E}_{p_\theta(y)}\left[\log p_\theta(y)\right].
\end{aligned}
$$

The discrete class probability $p_\theta(y)$ is estimated with samples $\mathbf{x}$ from the empirical model distribution according to $p_\theta(y) = \frac{1}{N}\sum_{i=1}^{N} p(y_i|\mathbf{x}_i)$ where $\mathbf{x}_i \sim p_\theta(\mathbf{x})$. The main limitation of the IS is that it is largely dependent on the performance of the classifier $f(\mathbf{x})$ on the dataset. Current practices use the InceptionNetv3 classifier trained on ImageNet.

The final metric we use to evaluate our models is the *Frechet inception distance*. Unlike the Inception score, the FID compares the synthesized samples directly with the true data samples. It does this by determining the discrepancy in statistics between feature vectors of the generated samples and data samples. This discrepancy is measured by the squared Wasserstein distance between two multidimensional Gaussian distributions and is given by:

$$\text{FID} = \|\mu - \mu_\theta\|^2 + \text{Tr}\left(\Sigma + \Sigma_\theta - 2\left(\Sigma\Sigma_\theta\right)^{1/2}\right) \tag{5.3}$$

where $\mathbf{z} \sim \mathcal{N}(\mu, \Sigma)$ and $\mathbf{z}_\theta \sim \mathcal{N}(\mu_\theta, \Sigma_\theta)$ are the gaussian distributions associated with the statistics of real and generated samples, respectively. In practice, these means and covariances are empirically computed based on the feature vectors from the InceptionNetv3 pool3 layer. Lower FID scores imply a smaller discrepancy between real and generated sample statistics and consequently better synthesis quality.

## 5.2 Results

This section details the qualitative and quantitative results of AdaNCE as a training algorithm. Performance is evaluated based on the synthesis quality of models trained on MNIST, CIFAR-10, and CelebA. Furthermore, each model is trained with several different $\mathcal{K}$ to analyze the effects of the adaptive interval on AdaNCE performance. Architecture A is used to train MNIST, while architecture C is used to train CIFAR-10 and CelebA.

### 5.2.1 Qualitative results

We notice that when the adaptive interval is $\mathcal{K} = 1$ corresponding to maximum likelihood learning, visual synthesis quality is comparable to the results of Du et al. (2019); Nijkamp et al. (2019a). However, when the adaptive interval increases the performance degrades. This is especially true of complex high-dimensional data like CIFAR and CelebA. In particular, there is a noticeable reduction in synthesis quality when $\mathcal{K} \approx 10$ for those aforementioned datasets.

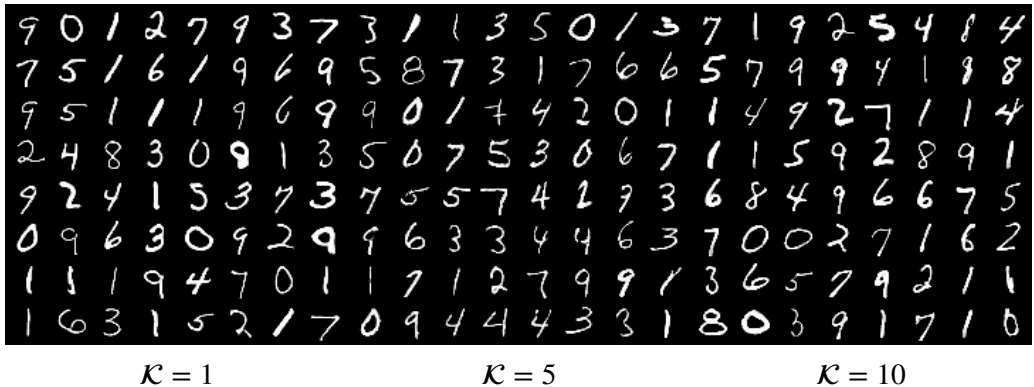

$\mathcal{K} = 1$        $\mathcal{K} = 5$        $\mathcal{K} = 10$

Figure 5.3   Short-run samples of EBMs trained on MNIST with self-adapting NCE. We show sample synthesis for a range of adaptive intervals $\mathcal{K}$.

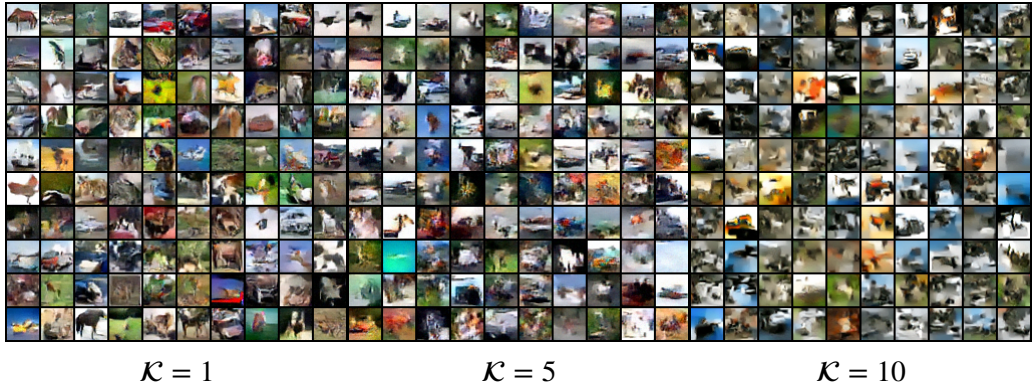

$\mathcal{K} = 1$          $\mathcal{K} = 5$          $\mathcal{K} = 10$

Figure 5.4   Short-run samples of EBMs trained on CIFAR-10 with self-adapting NCE. We show sample synthesis for a range of adaptive intervals $\mathcal{K}$.

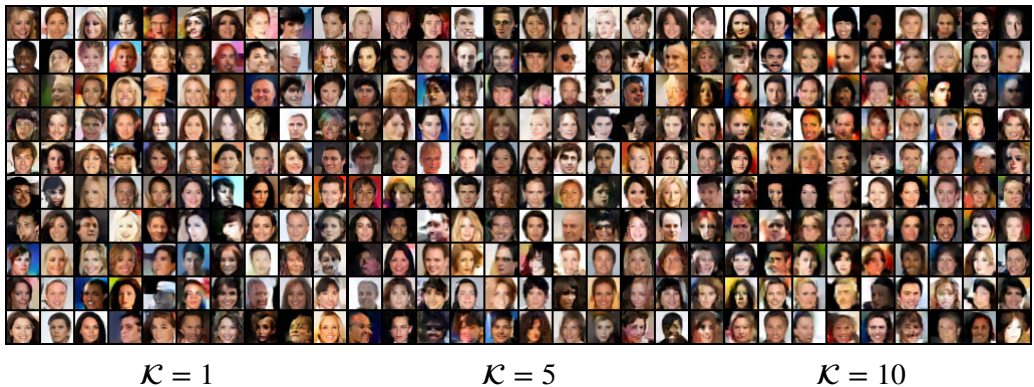

$\mathcal{K} = 1$          $\mathcal{K} = 5$          $\mathcal{K} = 10$

Figure 5.5   Short-run samples of EBMs trained on CelebA with self-adapting NCE. We show sample synthesis for a range of adaptive vintervals $\mathcal{K}$.

## 5.2.2   Quantitative results

We evaluate the Inception score and Frechet inception distance for models trained on CIFAR-10 with varying adaptive intervals $\mathcal{K}$. Unsurprisingly, we find that larger adaptive intervals coincide with worse synthesis quality, which agrees with our qualitative observations.

|     | $\mathcal{K} = 1$ | $\mathcal{K} = 5$ | $\mathcal{K} = 10$ | $\mathcal{K} = 20$ | $\mathcal{K} = 50$ |
|-----|-------|-------|-------|--------|--------|
| IS  | 4.03  | 3.74  | 3.57  | 3.45   | 2.61   |
| FID | 147.82 | 157.17 | 165.42 | 214.12 | 310.54 |

Table 5.2   IS and FID scores for models trained with varying adaptive intervals.

These evaluations compare unfavorably to the results of Du et al. (2019), whose strongest model recorded an Inception score of 6.02 and FID of 40.58. However, we note that their model utilized substantially deep residual networks, whereas our model relies on basic convolutional layers. In this regard, we emphasize that the purpose of this thesis is not to outperform existing architectures, but rather to provide insight between AdaNCE and maximum likelihood learning. In the case when $\mathcal{K} = 1$ our training algorithm is identical to that of maximum likelihood estimation.

## 5.3 Discussion

Both the qualitative and quantitative results show that the performance of AdaNCE degrades as the adaptive interval $\mathcal{K}$ increases. In the case when $\mathcal{K} = 1$ corresponding to maximum likelihood learning, AdaNCE performs similarly to results obtained by Nijkamp et al. (2019a). However, when the adaptive interval reaches about 10 steps per noise update, the synthesis quality deteriorates. This is especially true of complex high-dimensional data like CIFAR and CelebA. We provide two possible explanations for this phenomenon:

1. *Short-run MCMC introduces bias*: Since running Langevin dynamics to convergence is expensive, we instead choose to approximately sample from the self-adapting noise $p_{\mathrm{m}}(\mathbf{x})$ with short-run chains. However, these short-run chains introduce bias to the sampler, which may cause the learned EBM $p_\theta(\mathbf{x})$ to not approximate the data distribution $p_{\mathrm{data}}(\mathbf{x})$ well.

2. *The AdaNCE gradient estimator increases in variance*: Let us consider the gradient estimator of the AdaBRM objective (4.10) [1]:

$$\nabla_\theta L_{\mathrm{AdaBRM}}(\theta) = \mathbb{E}_{p_{\mathrm{m}}} \left[ S_0'(g_\theta) g_\theta \nabla_\theta \log \tilde{p}_\theta \right] - \mathbb{E}_{p_{\mathrm{data}}} \left[ S_1'(g_\theta) g_\theta \nabla_\theta \log \tilde{p}_\theta \right].$$

We note that the variance of this gradient estimator depends on the variance of the density ratio $g_\theta(\mathbf{x}) = \frac{\tilde{p}_\theta(\mathbf{x})}{\tilde{p}_{\mathrm{m}}(\mathbf{x})}$. As such, when the variance of the density ratio increases, the performance of AdaNCE degrades since its gradients can no longer be accurately estimated. In the case of maximum likelihood learning (i.e., $\mathcal{K} = 1$), this density ratio is constant and has zero variance, thus resulting in a gradient estimator with lower variance.

---

[1] Recall that in Chapter 4.3 we showed that AdaBRM is a generalization of AdaNCE.

# CHAPTER 6    CONCLUSIONS & FUTURE WORK

This thesis examined the task of learning energy-based models using noise-contrastive estimation. We first explained what noise-contrastive estimation is and how it can be used to train EBMs. While in theory, NCE is a powerful technique for probabilistic modelling, in practice it suffers greatly from the curse of dimensionality. Its performance is sensitive to the choice of the noise distribution $p_n(\mathbf{x})$, and is most effective when $p_n(\mathbf{x})$ approximates the data distribution. Several methods have been proposed that use an adaptive noise distribution to scale NCE to high-dimensional problems; however, they all suffer from several drawbacks. Conditional NCE (Ceylan et al., 2018) experiences vanishing gradients which prevent effective learning in high dimensions; while both adversarial contrastive estimation (Bose et al., 2018) and flow contrastive estimation (Gao et al., 2020) rely on adversarial objectives which can be challenging to train. This thesis attempted to improve noise-contrastive estimation with a self-adapting noise distribution that does not need to be adversarially trained. Furthermore, we uncovered a connection between our algorithm and maximum likelihood estimation through the framework of Bregman divergences. The key findings of this thesis are summarized below:

- *Self-adapting noise-contrastive estimation (AdaNCE)*: We presented a novel noise-contrastive estimation algorithm in which the noise distribution is taken to be static instances of the EBM along its training trajectory. This noise distribution progressively converges to the data distribution, thereby making NCE more scalable to high dimensions.

- *AdaNCE as a generalization of maximum likelihood learning*: We showed that the gradient estimators for both AdaNCE and maximum likelihood estimation are equivalent in the case where the adaptive interval is 1. Analogously, we also showed that density-ratio matching with Bregman divergences can also be reduced to maximum likelihood estimation.

- *Qualitative and quantitative evaluation of AdaNCE*: The synthesis quality of models trained with AdaNCE was evaluated. Results suggest that the performance of AdaNCE deteriorates as the adaptive intervals increase in length. This phenomenon is explained in the context of increased bias and variance.

Following this thesis, there are several possible directions for future work. One line of exploration involves mitigating the deteriorating performance of AdaNCE at larger adaptive intervals. This may include reducing the bias by using convergent MCMC rather than short-run MCMC. In addition, we can decrease the variance of the AdaNCE gradient estimators by introducing control variates. In particular, we may consider using Stein control variates which can theoretically result in complete variance reduction. Another potential direction involves exploring more efficient methods for sampling. Some authors (Nijkamp et al. (2019a)) have explored using Langevin dynamics without a noise term for approximate sampling. This is essentially equivalent to stochastic gradient descent. Building on this, we can consider adding a momentum term to Langevin dynamics for faster convergence. In the extreme case where we disregard the noise term, this becomes equivalent to sampling with SGD plus momentum. Lastly, we can consider the performance of self-adapting Bregman ratio-matching (AdaBRM) for a variety of convex functions $\Psi$. Certain choices of $\Psi$ lead to common statistical distances including the squared Mahalanobis distance and Kullback–Leibler divergence.

# APPENDIX A DERIVATIONS

## A.1 Denoising Score Matching

Let $p_\sigma(\mathbf{x}) = \int p_\sigma(\tilde{\mathbf{x}}|\mathbf{x}) p_{\text{data}}(\mathbf{x}) \, d\mathbf{x}$ be the Parzen density estimate of the data distribution $p_{\text{data}}(\mathbf{x})$ with gaussian kernel $p_\sigma(\tilde{\mathbf{x}}|\mathbf{x}) = \mathcal{N}(\tilde{\mathbf{x}}; \mathbf{x}, \sigma^2 I)$. The explicit score matching objective between the Parzen estimate of the data distribution and model distribution is

$$
\begin{aligned}
J_{\text{DSM}}(\theta) &= \mathbb{E}_{p_\sigma(\tilde{\mathbf{x}})} \left[ \frac{1}{2} \| \nabla_{\tilde{\mathbf{x}}} \log p_\sigma(\tilde{\mathbf{x}}) - \nabla_{\tilde{\mathbf{x}}} \log p_\theta(\tilde{\mathbf{x}}) \|^2 \right] \\
&= \mathbb{E}_{p_\sigma(\tilde{\mathbf{x}})} \left[ \frac{1}{2} \| \nabla_{\tilde{\mathbf{x}}} \log p_\sigma(\tilde{\mathbf{x}}) \|^2 + \frac{1}{2} \| \nabla_{\tilde{\mathbf{x}}} \log p_\theta(\tilde{\mathbf{x}}) \|^2 - \nabla_{\tilde{\mathbf{x}}} \log p_\sigma(\tilde{\mathbf{x}})^\mathsf{T} \nabla_{\tilde{\mathbf{x}}} \log p_\theta(\tilde{\mathbf{x}}) \right] \\
&= \mathbb{E}_{p_\sigma(\tilde{\mathbf{x}})} \left[ \frac{1}{2} \| \nabla_{\tilde{\mathbf{x}}} \log p_\theta(\tilde{\mathbf{x}}) \|^2 - \nabla_{\tilde{\mathbf{x}}} \log p_\sigma(\tilde{\mathbf{x}})^\mathsf{T} \nabla_{\tilde{\mathbf{x}}} \log p_\theta(\tilde{\mathbf{x}}) \right] + \text{const.} \\
&= \mathbb{E}_{p_\sigma(\tilde{\mathbf{x}})} \left[ \frac{1}{2} \| \nabla_{\tilde{\mathbf{x}}} \log p_\theta(\tilde{\mathbf{x}}) \|^2 \right] - \int p_\sigma(\tilde{\mathbf{x}}) \nabla_{\tilde{\mathbf{x}}} \log p_\sigma(\tilde{\mathbf{x}})^\mathsf{T} \nabla_{\tilde{\mathbf{x}}} \log p_\theta(\tilde{\mathbf{x}}) \, d\tilde{\mathbf{x}} + \text{const.} \\
&= \mathbb{E}_{p_\sigma(\tilde{\mathbf{x}})} \left[ \frac{1}{2} \| \nabla_{\tilde{\mathbf{x}}} \log p_\theta(\tilde{\mathbf{x}}) \|^2 \right] - \int \nabla_{\tilde{\mathbf{x}}} p_\sigma(\tilde{\mathbf{x}})^\mathsf{T} \nabla_{\tilde{\mathbf{x}}} \log p_\theta(\tilde{\mathbf{x}}) \, d\tilde{\mathbf{x}} + \text{const.}
\end{aligned}
$$

We substitute the expression for the Parzen window estimate and then manipulate the expression as follows:

$$
\begin{aligned}
\int \nabla_{\tilde{\mathbf{x}}} p_\sigma(\tilde{\mathbf{x}})^\mathsf{T} \nabla_{\tilde{\mathbf{x}}} \log p_\theta(\tilde{\mathbf{x}}) \, d\tilde{\mathbf{x}} &= \int \nabla_{\tilde{\mathbf{x}}} \left( \int p_\sigma(\tilde{\mathbf{x}}|\mathbf{x}) p_{\text{data}}(\mathbf{x}) \, d\mathbf{x} \right)^\mathsf{T} \nabla_{\tilde{\mathbf{x}}} \log p_\theta(\tilde{\mathbf{x}}) \, d\tilde{\mathbf{x}} \\
&= \int \nabla_{\tilde{\mathbf{x}}} p_\sigma(\tilde{\mathbf{x}}|\mathbf{x})^\mathsf{T} \nabla_{\tilde{\mathbf{x}}} \log p_\theta(\tilde{\mathbf{x}}) p_{\text{data}}(\mathbf{x}) \, d\mathbf{x} \, d\tilde{\mathbf{x}} \\
&= \int \nabla_{\tilde{\mathbf{x}}} \log p_\sigma(\tilde{\mathbf{x}}|\mathbf{x})^\mathsf{T} \nabla_{\tilde{\mathbf{x}}} \log p_\theta(\tilde{\mathbf{x}}) p_\sigma(\tilde{\mathbf{x}}|\mathbf{x}) p_{\text{data}}(\mathbf{x}) \, d\mathbf{x} \, d\tilde{\mathbf{x}} \\
&= \mathbb{E}_{p_\sigma(\tilde{\mathbf{x}}, \mathbf{x})} \left[ \nabla_{\tilde{\mathbf{x}}} \log p_\sigma(\tilde{\mathbf{x}}|\mathbf{x})^\mathsf{T} \nabla_{\tilde{\mathbf{x}}} \log p_\theta(\tilde{\mathbf{x}}) \right].
\end{aligned}
$$

Substituting this expression back into the original objective results in denoising score matching:

$$
\begin{aligned}
J_{\text{DSM}}(\theta) &= \mathbb{E}_{p_\sigma(\tilde{\mathbf{x}}, \mathbf{x})} \left[ \frac{1}{2} \| \nabla_{\tilde{\mathbf{x}}} \log p_\theta(\tilde{\mathbf{x}}) \|^2 - \nabla_{\tilde{\mathbf{x}}} \log p_\sigma(\tilde{\mathbf{x}}|\mathbf{x})^\mathsf{T} \nabla_{\tilde{\mathbf{x}}} \log p_\theta(\tilde{\mathbf{x}}) \right] + \text{const.} \\
&= \mathbb{E}_{p_\sigma(\tilde{\mathbf{x}}, \mathbf{x})} \left[ \frac{1}{2} \| \nabla_{\tilde{\mathbf{x}}} \log p_\theta(\tilde{\mathbf{x}}) \|^2 - \nabla_{\tilde{\mathbf{x}}} \log p_\sigma(\tilde{\mathbf{x}}|\mathbf{x})^\mathsf{T} \nabla_{\tilde{\mathbf{x}}} \log p_\theta(\tilde{\mathbf{x}}) \right. \\
&\qquad\qquad \left. + \frac{1}{2} \| \nabla_{\tilde{\mathbf{x}}} \log p_\sigma(\tilde{\mathbf{x}}|\mathbf{x}) \|^2 \right] + \text{const.} \\
&= \mathbb{E}_{p_\sigma(\tilde{\mathbf{x}}, \mathbf{x})} \left[ \frac{1}{2} \| \nabla_{\tilde{\mathbf{x}}} \log p_\sigma(\tilde{\mathbf{x}}|\mathbf{x}) - \nabla_{\tilde{\mathbf{x}}} \log p_\theta(\tilde{\mathbf{x}}) \|^2 \right] + \text{const.},
\end{aligned}
$$

where in the second line we can add the inner product term since it does not depend on the model parameters $\theta$ (i.e., it is a constant).

## A.2 SSM and Hutchinson Equivalence

Assuming that $\mathbf{v}$ has mean 0 and covariance $I$ then the squared scalar projection term can be written as:

$$
\begin{aligned}
\mathbb{E}_{\mathbf{v}}\left[\left(\mathbf{v}^{\mathsf{T}}\nabla_{\mathbf{x}}\log p_{\theta}(\mathbf{x})\right)^{2}\right] &= \mathbb{E}_{\mathbf{v}}\left[\mathbf{v}^{\mathsf{T}}\nabla_{\mathbf{x}}\log p_{\theta}(\mathbf{x})\nabla_{\mathbf{x}}\log p_{\theta}(\mathbf{x})^{\mathsf{T}}\mathbf{v}\right] \\
&= \mathrm{Tr}\left(\nabla_{\mathbf{x}}\log p_{\theta}(\mathbf{x})\nabla_{\mathbf{x}}\log p_{\theta}(\mathbf{x})^{\mathsf{T}}\right) \\
&= \nabla_{\mathbf{x}}\log p_{\theta}(\mathbf{x})^{\mathsf{T}}\nabla_{\mathbf{x}}\log p_{\theta}(\mathbf{x}) \\
&= \|\nabla_{\mathbf{x}}\log p_{\theta}(\mathbf{x})\|^{2},
\end{aligned}
$$

where the second line is a result of $\mathbb{E}[\mathbf{v}^{\mathsf{T}}A\mathbf{v}] = \mathrm{Tr}(A)$ for any matrix A, and the third line is due to the cyclic property of trace (a.k.a., trace trick).

# ACKNOWLEDGEMENTS

I would like to express my sincere gratitude to my supervisor, Professor Zhu, for his careful guidance and to the members of the TSAIL lab for their thought provoking discussions and support. I started this Masters program knowing barely anything about machine learning, and I am forever grateful for the amount of knowledge that my supervisor has inspired me to learn. In particular, I greatly appreciate the research freedom that was afforded to me by Professor Zhu at TSAIL. Without the opportunity to explore a variety of different directions and ideas, I would not be the researcher I am today.

I want to also thank my parents for always being there for me in my moments of failure and doubt. While we may disagree at times, I am forever grateful for their love and support. Without them, I would not be the person I am today.

To my closest friends Matt, Andrew, and Camilla, thank you for all the entertaining stories, dinners, and gaming nights. Studying at Tsinghua would not be the experience it is without you guys. While this chapter of my life is over, I am sure we will see each other again in the future.

Finally, I want to thank Sunny. Thank you for supporting me throughout these years, and for being that person I can always talk to at the end of the day. You are what motivated me to keep going.